\def\year{2019}\relax
\documentclass[letterpaper]{article} %DO NOT CHANGE THIS
\usepackage{aaai19}  %Required
\usepackage{times}  %Required
\usepackage{helvet}  %Required
\usepackage{courier}  %Required
\usepackage{url}  %Required
\usepackage{graphicx}  %Required

\PassOptionsToPackage{numbers, compress}{natbib}

\usepackage{mathtools}
\usepackage{amsmath}
\usepackage{amsfonts}
\usepackage{dsfont}
\usepackage{bbm}
\usepackage{amsthm}
\usepackage{enumerate}

\newcommand{\bb}{\boldsymbol{b}}

\newcommand{\eb}{\boldsymbol{e}}

\newcommand{\hb}{\boldsymbol{h}}

\newcommand{\wb}{\boldsymbol{w}}

\newcommand{\Wb}{\boldsymbol{W}}

\newcommand{\Expectation}{\mathbb{E}}

\newtheorem{property}{Property}

\begin{document}

\title{EA-CG: An Approximate Second-Order Method \\ for Training Fully-Connected Neural Networks}

\author{
  Sheng-Wei Chen\\
  HTC Research \& Healthcare\\
  sw\_chen@htc.com \\
  \And
  Chun-Nan Chou\\
  HTC Research \& Healthcare\\
  jason.cn\_chou@htc.com\\
  \And
  Edward Y. Chang\\
  HTC Research \& Healthcare\\
  edward\_chang@htc.com\\
}
\maketitle

\begin{abstract}
For training fully-connected neural networks (FCNNs), we propose a practical approximate second-order method including: 1) an approximation of the Hessian matrix and 2) a conjugate gradient (CG) based method.
Our proposed approximate Hessian matrix is memory-efficient and can be applied to any FCNNs where the activation and criterion functions are twice differentiable. 
%We also devise an effective CG-based method to derive Newton directions for training FCNNs in mini-batch setting.
%Our devised CG-based method reduces the space and time complexity of the CG method significantly and can be employed in any approximate Hessian matrix of FCNNs that is positive semi-definite.
We devise a CG-based method incorporating one-rank approximation to derive Newton directions for training FCNNs, which significantly reduces both space and time complexity.
This CG-based method can be employed to solve any linear equation where the coefficient matrix is Kronecker-factored, symmetric and positive definite.
Empirical studies show the efficacy and efficiency of our proposed method. 
\end{abstract}

\section{Introduction}
\label{sec:Intro}

Neural networks have been applied to solving problems in several application domains such as computer vision~\cite{KH16a}, natural language processing~\cite{SH97a}, and disease diagnosis~\cite{EYC17a}. %XSW17a,
Training a neural network requires tuning its model parameters using Backpropagation. % \cite{PJW74a}. 
Stochastic gradient descent (SGD), Broyden-Fletcher-Goldfarb-Shanno and one-step secant are representative algorithms that have been employed for training in Backpropagation. 

To date, SGD is widely used due to its low computational demand. 
SGD minimizes a function using the function's first derivative, and has been proven to be effective for training large models.
However, stochasticity in the gradient will slow down convergence for any gradient method such that none of them can be asymptotically faster than simple SGD with Polyak averaging~\cite{BP92a}. % .
Besides gradients, second-order methods utilize the curvature information of a loss function within the neighborhood of a given point to guide the update direction.
Since each update becomes more precise, such methods can converge faster than first-order methods in terms of update iterations.

For solving a convex optimization problem, a second-order method can always converge to the global minimum in much fewer steps than SGD. 
However, the problem of neural-network training can be non-convex, thereby suffering from the issue of negative curvature.
To avoid this issue, the common practice is to use the Gauss-Newton matrix with a convex criterion function~\cite{NS02a} or the Fisher matrix to measure curvature since both are guaranteed to be positive semi-definite (PSD). 

Although these two kinds of matrices can alleviate the issue of negative curvature, computing either the exact Gauss-Newton matrix or Fisher matrix even for a modestly-sized fully-connected neural network (FCNN) is intractable.
Intuitively, the analytic expression for the second derivative requires $O(N^2)$ computations if $O(N)$ complexity is required to compute the first derivative. 
Thus, several pioneer works~\cite{LY98a,SA00a,NS02a} have used different methods to approximate either matrix. 
However, none of these methods have been shown to be computationally feasible and fundamentally more effective than first-order methods as reported in \cite{JM10a}.
Thus, there has been a growing trend towards conceiving more computationally feasible second-order methods for training FCNNs.

We outline several notable works in chronological order herein.
\citeauthor{JM10a} proposed a truncated-Newton method for training deep auto-encoders. % \cite{SJW99a}
In this work, \citeauthor{JM10a} used an $\mathcal{R}$-operator~\cite{BAP94a} to compute full Gauss-Newton matrix-vector products and made good progress within each update. 
\citeauthor{JM15a} developed a block-diagonal approximation to the Fisher matrix for FCNNs, called Kronecker-factored Approximation Curvature (KFAC).
They derived the update directions by exploiting the inverse property of Kronecker products.
KFAC features that its cost in storing and inverting the devised approximation does not depend on the amount of data used to estimate the Fisher matrix.
The idea is further extended to convolutional nets~\cite{RG16a}.
Recently, \citeauthor{AB17a} presented a block-diagonal approximation to the Gauss-Newton matrix for FCNNs, referred to as Kronecker-Factored Recursive Approximation (KFRA).
They also utilized the inverse property of Kronecker products to derive the update directions.
Similarly, \citeauthor{HZ17a} introduced a block-diagonal approximation of the Gauss-Newton matrix and used conjugate gradient (CG) method to derive the update directions. %\cite{SJW99a} 

However, these prior works either impose constraints on their applicability or require considerable computation in terms of memory or running time.
On the one hand, several of these notable methods relying on the Gauss-Newton matrix face the essential limit that they cannot handle non-convex criterion functions playing an important part in some problems.
Use robust estimation in computer vision \cite{CS99a} as an example. 
Some non-convex criterion functions such as Tukey’s biweight function are robustness to outliers and perform better than convex criterion functions \cite{VB15a}.
On the other hand, in order to derive the update directions, some of the methods utilizing the conventional CG method may take too much time, and the others exploiting the inverse property of Kronecker products may require excessive memory space.

To remedy the aforementioned issues, we propose a block-diagonal approximation of the positive-curvature Hessian (PCH) matrix, which is memory-efficient. 
Our proposed PCH matrix can be applied to any FCNN where the activation and criterion functions are twice differentiable.
Particularly, our proposed PCH matrix can handle non-convex criterion functions, which the Gauss-Newton methods cannot. 
Besides, we incorporate expectation approximation into the CG-based method, which is dubbed EA-CG, to derive update directions for training FCNNs in mini-batch setting.
EA-CG significantly reduces the space and time complexity of the conventional CG method.
Our experimental results show the efficacy and efficiency of our proposed method. 

In this work, we focus on deriving a second-order method for training FCNNs since the shared weights of convolutional layers lead to the difficulties in factorizing its Hessian. 
We defer tackling convolutional layers to our future work.
We also focus on the classification problem and hence do not consider auto-encoders.
Our strategy is that once a simpler isolated problem can be effectively handled, we can then extend our method to address more challenging issues.

In summary, the contributions of this paper are as follows: 
\begin{enumerate}
  \item For curvature information, we propose the PCH matrix to improve the Gauss-Newton matrix for training FCNNs with convex criterion functions and overcome the non-convex scenario. 

  \item To derive the update directions, we devise effective EA-CG method, which does converge faster in terms of wall clock time and enjoys better testing accuracy than competing methods. Specially, the performance of EA-CG is competitive with SGD.  
\end{enumerate}

\section{Truncated-Newton Method on Non-Convex Problems}
\label{sec:Truncated_Newton_Method}

Newton's method is one of the second-order minimization methods, and is generally composed of two steps: 1) computing the Hessian matrix and 2) solving the system of linear equations for update directions.
The truncated-Newton method applies the CG method with restricted iterations to the second step of Newton's method.
In this section, we first introduce the truncated-Newton method in the context of convex problems.
Afterwards, we discuss the non-convex scenario of the truncated-Newton method and provide an important property that lays the foundation of our proposed PCH matrix. 

%Separate into two subsections?
%\subsection{Convex Problem Scenario}
Suppose we have a minimization problem
\begin{equation}
  \label{eq:Newton_original}
  \min_{\boldsymbol{\theta}} f(\boldsymbol{\theta}),
\end{equation}
where $f$ is a convex and twice-differentiable function. 
Since the global minimum is at the point that the first derivative is zero, the solution $\boldsymbol{\theta}^{*}$ can be derived from the equation
\begin{equation}
  \label{eq:Gradient_Equal_Zero}
  \nabla f(\boldsymbol{\theta}^{*}) = 0.
\end{equation}
We can utilize a quadratic polynomial to approximate Problem \ref{eq:Newton_original} by conducting a Taylor expansion with a given point $\boldsymbol{\theta}^{j}$.
Then, the problem turns out to be 
\[
  \min_{\boldsymbol{d}} f(\boldsymbol{\theta}^{j} + \boldsymbol{d}) \approx f(\boldsymbol{\theta}^{j}) + \nabla f(\boldsymbol{\theta}^{j})^{T} \boldsymbol{d} + \frac{1}{2} \boldsymbol{d}^{T} \nabla^{2}f(\boldsymbol{\theta}^{j}) \boldsymbol{d},
\]
where $\nabla^{2}f(\boldsymbol{\theta}^{j})$ is the Hessian matrix of $f$ at $\boldsymbol{\theta}^{j}$.
After applying the aforementioned approximation, we rewrite Eq.~(\ref{eq:Gradient_Equal_Zero}) as the linear equation
\begin{equation}
  \label{eq:Newton_Direction_Linear_Equation}
  \nabla f(\boldsymbol{\theta}^{j}) + \nabla^{2}f(\boldsymbol{\theta}^{j}) \boldsymbol{d}^{j} = 0.
\end{equation}
Therefore, the Newton direction is obtained via 
\[
  \boldsymbol{d}^{j} = - \nabla^{2}f(\boldsymbol{\theta}^{j})^{-1} \nabla f(\boldsymbol{\theta}^{j}),
\]
and we can acquire $\boldsymbol{\theta}^{*}$ by iteratively applying the update rule
\[
  \boldsymbol{\theta}^{j+1} = \boldsymbol{\theta}^{j} + \eta \boldsymbol{d}^{j},
\]
where $\eta$ is the step size.

%\subsection{Non-Convex Problem Scenario}
For non-convex problems, the solution to Eq.~(\ref{eq:Gradient_Equal_Zero}) reflects one of three possibilities: a local minimum $\boldsymbol{\theta}_{\text{min}}$, a local maximum $\boldsymbol{\theta}_{\text{max}}$ or a saddle point $\boldsymbol{\theta}_{\text{saddle}}$.
Some previous works such as \cite{YND14a,DG17a,ME08a} utilize negative curvature information to converge to a local minimum.
Before illustrating how these previous works tackle the issue of negative curvature, we have to introduce a crucial concept that we can know the curvature information of $f$ at a given point $\boldsymbol{\theta}$ by analyzing the Hessian matrix $\nabla^{2}f(\boldsymbol{\theta})$.
On the one hand, the Hessian matrix of $f$ at any $\boldsymbol{\theta}_{\text{min}}$ is positive semi-definite.
On the other hand, the Hessian matrices of $f$ at any $\boldsymbol{\theta}_{\text{max}}$ and $\boldsymbol{\theta}_{\text{saddle}}$ are negative semi-definite and indefinite, respectively. 
After establishing the concept, we can use the following property to understand how to utilize the negative curvature information to resolve the issue of negative curvature.
\begin{property}
\label{prop:Negative_Eigenvalues_Implies_Concave_Part}
Let $f$ be a non-convex and twice-differentiable function.
With a given point $\boldsymbol{\theta}^{j}$, we suppose that there exist some negative eigenvalues $\{\lambda_{1}, \ldots, \lambda_{s}\}$ for $\nabla^{2}f(\boldsymbol{\theta}^{j})$.
Moreover, we take $V = \text{span}(\{\boldsymbol{v}_{1}, \ldots, \boldsymbol{v}_{s} \})$, which is the eigenspace corresponds to $\{\lambda_{1}, \ldots, \lambda_{s}\}$.
If we take 
\[ 
  g(\boldsymbol{k}) = f(\boldsymbol{\theta}^{j}) + {\nabla f(\boldsymbol{\theta}^{j})}^{T} \boldsymbol{v} + \frac{1}{2} \boldsymbol{v}^{T} \nabla^{2} f(\boldsymbol{\theta}^{j}) \boldsymbol{v} ,
\]
where $\boldsymbol{k} \in \mathbb{R}^{s} \text{ and } \boldsymbol{v} = k_{1}\boldsymbol{v}_{1} + \ldots + k_{s}\boldsymbol{v}_{s}$, then $g(\boldsymbol{k})$ is a concave function.
\end{property}

According to \textbf{Property~\ref{prop:Negative_Eigenvalues_Implies_Concave_Part}}, Eq.~(\ref{eq:Newton_Direction_Linear_Equation}) may lead us to a local maximum or a saddle point if $\nabla^{2} f(\boldsymbol{\theta}^{j})$ has some negative eigenvalues.
In order to converge to a local minimum, we substitute $\text{Pos-Eig}(\nabla^{2}f(\boldsymbol{\theta}^{j}))$ for $\nabla^{2}f(\boldsymbol{\theta}^{j})$, where $\text{Pos-Eig}(A)$ is conceptually defined as replacing the negative eigenvalues of $A$ with non-negative ones.
That is,
\[
  \text{Pos-Eig}(A) = Q^{T} 
    \begin{bmatrix} \gamma \lambda_{1} &        &                    &                &        &             \\
                                       & \ddots &                    &                &        &             \\
                                       &        & \gamma \lambda_{s} &                &        &             \\
                                       &        &                    & \lambda_{s+1}  &        &             \\
                                       &        &                    &                & \ddots &             \\
                                       &        &                    &                &        & \lambda_{n} \\
    \end{bmatrix} 
    Q,
\]
where $\gamma$ is a given scalar that is less than or equal to zero, and $\{\lambda_{1}, \ldots, \lambda_{s}\}$ and $\{\lambda_{s+1}, \ldots, \lambda_{n}\}$ are the negative and non-negative eigenvalues of $A$, respectively.
This refinement implies that the point $\boldsymbol{\theta}^{j+1}$ escapes from either local maxima or saddle points if $\gamma < 0$.
In case of $\gamma = 0$, this refinement means that the eigenspace of the negative eigenvalues is ignored.
As a result, we do not converge to any saddle point or local maximum.
In addition, every real symmetric matrix can be diagonalized according to the spectral theorem.
Under our assumptions, $\nabla^{2}f(\boldsymbol{\theta}^{j})$ is a real symmetric matrix.
Thus, $\nabla^{2}f(\boldsymbol{\theta}^{j})$ can be decomposed, and the function ``$\text{Pos-Eig}$'' can be realized easily. 

When the number of variables in $f$ is large, the Hessian matrix becomes intractable with respect to space complexity.
Alternatively, we can utilize the CG method to solve Eq.~(\ref{eq:Newton_Direction_Linear_Equation}).
This alternative only requires calculating the Hessian-vector products rather than storing the whole Hessian matrix.
Moreover, to save computation costs, it is desirable to restrict the iteration number of the CG method. 

\section{Computing the Hessian Matrix}
\label{sec:Backpropagation}

For second-order methods we must compute the curvature information, and we utilize the Hessian matrix to capture the curvature information in our work.
However, the Hessian matrix for training FCNNs is intrinsically complicated and intractable. 
Recently, \citeauthor{AB17a} presented the idea of block Hessian recursion, in which the diagonal blocks of the Hessian matrix can be computed in a layer-wise manner.
As the basis of our proposed PCH matrix, we first establish some notation for training FCNNs and reformulate the block Hessian recursion with our notation. 
Then, we present the steps to integrate the approximation concept proposed by \cite{JM15a} into our reformulation.
%Full derivations are provided in the supplementary material. 

\subsection{Fully-Connected Neural Networks}
\label{sec:FCNN}

An FCNN with $k$ layers takes an input vector $\boldsymbol{h}_{i}^{0} = \boldsymbol{x}_{i}$, where $\boldsymbol{x}_{i}$ is the $i^{\text{th}}$ instance in the training set.
For the $i^{\text{th}}$ instance, the activation values in the other layers can be recursively derived from: $\boldsymbol{h}_{i}^{t} = \sigma(\boldsymbol{W}^{t}\boldsymbol{h}_{i}^{t-1} + \boldsymbol{b}^{t}), t = 1, \ldots, k-1,$ where $\sigma$ is the activation function and can be any twice differentiable function, and $\boldsymbol{W}^{t}$ and $\boldsymbol{b}^{t}$ are the weights and biases in the $t^{\text{th}}$ layer, respectively.
We further denote $n_{t}$ as the number of the neurons in the $t^{\text{th}}$ layer, where $t = 0, \ldots, k$, and collect all the model parameters including all the weights and biases in each layer as $\boldsymbol{\theta} = (\text{Vec}(\boldsymbol{W}^{1}), \boldsymbol{b}^{1}, \ldots, \text{Vec}(\boldsymbol{W}^{k}), \boldsymbol{b}^{k} ),$ where $\text{Vec}(A) = \begin{bmatrix} [A_{\cdot 1}]^{T} & [A_{\cdot 2}]^{T} & \cdots & [A_{\cdot n}]^{T}  \end{bmatrix}^{T}$.
By following the notation mentioned above, we denote an FCNN output with $k$ layers as $\boldsymbol{h}_{i}^{k} = F( \boldsymbol{\theta} |\boldsymbol{x}_{i}) = \boldsymbol{W}^{k}\boldsymbol{h}_{i}^{k-1} + \boldsymbol{b}^{k}$.

To train this FCNN, we must decide a loss function $\xi$ that can be any twice differentiable function. 
Training this FCNN can therefore be interpreted as solving the following minimization problem:
\begin{equation*}
  \min_{\boldsymbol{\theta}} \sum_{i=1}^{l} \xi(\boldsymbol{h}_{i}^{k} \mid \boldsymbol{y}_{i}) \equiv \min_{\boldsymbol{\theta}} \sum_{i=1}^{l} C( \hat{\boldsymbol{y}}_{i} \mid \boldsymbol{y}_{i}),
\end{equation*}
where $l$ is the number of the instances in the training set, $\boldsymbol{y}_{i}$ is the label of the $i^{\text{th}}$ instance, $\hat{\boldsymbol{y}}_{i}$ is $\text{softmax}(\boldsymbol{h}_{i}^{k})$, and $C$ is the criterion function.

\subsection{Layer-wise Equations for the Hessian Matrix}
\label{sec:EquationHessianInfo}

For lucid exposition of the block Hessian recursion, we start by reformulating the equations of Backpropagation according to the notation defined in the previous subsection.
Please note that we separate the bias terms ($\boldsymbol{b}^{t}$) from the weight terms ($\boldsymbol{W}^{t}$) and treat each of them individually during backward propagation of gradients.
The gradients of $\xi$ with respect to the bias and weight terms can be derived from our reformulated equations in a layer-wise manner, similar to the original Backpropagation method.
For the $i^{\text{th}}$ instance, our reformulated equations are as follows:

\begin{equation*}
\begin{split}
  \nabla_{\boldsymbol{b}^{k}}\xi_{i}   =& \nabla_{\boldsymbol{h}_{i}^{k}}\xi_{i} \\
  \nabla_{\boldsymbol{b}^{t-1}}\xi_{i} =& \text{diag}(\boldsymbol{h}_{i}^{(t-1)'}) \boldsymbol{W}^{tT} \nabla_{\boldsymbol{b}^{t}}\xi_{i} \\
  \nabla_{\boldsymbol{W}^{t}}\xi_{i}   =& \nabla_{\boldsymbol{b}^{t}}\xi_{i} \otimes \boldsymbol{h}_{i}^{(t-1)T} \\
\end{split}
\end{equation*}
where $\xi_{i} = \xi(\boldsymbol{h}_{i}^{k} \mid \boldsymbol{y}_{i})$, $\otimes$ is the Kronecker product, and $\boldsymbol{h}_{i}^{(t-1)'} = \left. \nabla_{\boldsymbol{z}}\sigma(\boldsymbol{z}) \right|_{\boldsymbol{z} = \boldsymbol{W}^{t-1}\boldsymbol{h}_{i}^{t-2} + \boldsymbol{b}^{t-1} }$.
Likewise, we strive to propagate the Hessian matrix of $\xi$ with respect to the bias and weight terms backward in a layer-wise manner.
This can be achieved by utilizing the Kronecker product and following the similar fashion above.
The resultant equations for the $i^{\text{th}}$ instance are as follows:
\begin{subequations}
\begin{align}
  \nabla_{\boldsymbol{b}^{k}}^{2} \xi_{i}   =& \nabla_{\boldsymbol{h}_{i}^{k}}^{2} \xi_{i} \\
  \nabla_{\boldsymbol{b}^{t-1}}^{2} \xi_{i} =& \text{diag}(\boldsymbol{h}_{i}^{(t-1)'}) \boldsymbol{W}^{tT} \nabla_{\boldsymbol{b}^{t}}^{2} \xi_{i} \boldsymbol{W}^{t} \text{diag}(\boldsymbol{h}_{i}^{(t-1)'}) \nonumber \\
                                  & + \text{diag}(\boldsymbol{h}_{i}^{(t-1)''}  \odot (\boldsymbol{W}^{tT} \nabla_{\boldsymbol{b}^{t}} \xi_{i})) \label{eq:PropagateHessian} \\
  \nabla_{\boldsymbol{W}^{t}}^{2} \xi_{i}   =& ( \boldsymbol{h}_{i}^{t-1} \otimes \boldsymbol{h}_{i}^{(t-1)T} ) \otimes \nabla_{\boldsymbol{b}^{t}}^{2} \xi_{i}, \label{eq:PropagateHessian_W}
\end{align}
\end{subequations}
where $\odot$ is the element-wise product, $\left[ \boldsymbol{h}_{i}^{(t-1)''}\right]_{s} = \left[ \left. \nabla_{\boldsymbol{z}}^{2} \sigma(\boldsymbol{z}) \right|_{\boldsymbol{z} = \boldsymbol{W}^{t-1}\boldsymbol{h}_{i}^{t-2} + \boldsymbol{b}^{t-1} } \right]_{ss}$, and the derivative order of $\nabla_{\boldsymbol{W}^{t}}^{2} \xi_{i}$ is column-wise traversal of $\boldsymbol{W}^{t}$. 
The derivations are in the Supplementary Materials B and C.
Moreover, it is worth noting that the original block Hessian recursion unified the bias and weight terms, which is distinct from our separate treatment of these terms. 

\subsection{Expectation Approximation}
\label{sec:Expectation_Approximation}

\citeauthor{JM15a} propose one approximation concept that is referred as {\em expectation approximation} in \cite{AB17a}. 
The idea behind expectation approximation is that the covariance between $[ \boldsymbol{h}_{i}^{t-1} \otimes \boldsymbol{h}_{i}^{(t-1)T} ]_{uv}$ and $[\nabla_{\boldsymbol{b}^{t}} \xi_{i} \otimes {\nabla_{\boldsymbol{b}^{t}} \xi_{i}}^{T} ]_{\mu \nu} $ with given indices $(u, v)$ and $(\mu, \nu)$ is shown to be tiny and thus ignored due to computational efficiency, i.e.,
\[
\begin{split}
         & \mathbb{E}_{i}[ [\boldsymbol{h}_{i}^{t-1} \otimes \boldsymbol{h}_{i}^{(t-1)T}]_{uv} \cdot [\nabla_{\boldsymbol{b}^{t}}\xi_{i} \otimes {\nabla_{\boldsymbol{b}^{t}}\xi_{i}}^{T}]_{\mu \nu} ] \\
  \approx& \mathbb{E}_{i}[ [\boldsymbol{h}_{i}^{t-1} \otimes \boldsymbol{h}_{i}^{(t-1)T}]_{uv} ] \cdot \mathbb{E}_{i}[ [\nabla_{\boldsymbol{b}^{t}}\xi_{i} \otimes {\nabla_{\boldsymbol{b}^{t}}\xi_{i}}^{T}]_{\mu \nu} ]. \\
\end{split}
\]
To explain this concept on our formulations, we define cov-$t$ as $\text{Ele-Cov}(( \boldsymbol{h}_{i}^{t-1} \otimes \boldsymbol{h}_{i}^{(t-1)T} ) \otimes \mathds{1}_{ n_{t}, n_{t} }, \mathds{1}_{ n_{t-1}, n_{t-1} } \otimes \nabla_{\boldsymbol{b}^{t}}^{2} \xi_{i} )$, where "Ele-Cov" is denoted as element-wise covariance, and $\mathds{1}_{u,v}$ is the matrix whose elements are $1$ in $\mathbb{R}^{u \times v}$, $t = 1, \ldots, k$. 
With the definition of cov-$t$ and our devised equations in the previous subsection, the approximation can be interpreted as follows:
\begin{align}
  \mathbb{E}_{i} [\nabla_{\boldsymbol{W}^{t}}^{2} \xi_{i}] =& \text{EhhT}^{t-1}
                                                     \otimes \mathbb{E}_{i}[ \nabla_{\boldsymbol{b}^{t}}^{2} \xi_{i}] + \text{cov-${t}$} \nonumber \\
                                            \approx& \text{EhhT}^{t-1}
                                                     \otimes \mathbb{E}_{i}[ \nabla_{\boldsymbol{b}^{t}}^{2} \xi_{i}], \label{eq:PropagateHessian_W_mathbb{E}} 
\end{align}
where $\text{EhhT}^{t-1} = \mathbb{E}_{i} [ \boldsymbol{h}_{i}^{t-1} \otimes \boldsymbol{h}_{i}^{(t-1)T} ] $. 

\citeauthor{AB17a} also adopted expectation approximation in their proposed method. 
Similarly, we integrate this approximation into our proposed layer-wise Hessian matrix equations, thereby resulting in the following approximation equation:
\begin{align}
         & \mathbb{E}_{i}[\nabla_{\boldsymbol{b}^{t-1}}^{2} \xi_{i}] \nonumber \\ 
  \approx& \mathbb{E}_{i}[\text{diag}(\boldsymbol{h}_{i}^{(t-1)'}) \boldsymbol{W}^{tT} \mathbb{E}_{i}[ \nabla_{\boldsymbol{b}^{t}}^{2} \xi_{i}] \boldsymbol{W}^{t} \text{diag}(\boldsymbol{h}_{i}^{(t-1)'}) \nonumber \\
         & + \text{diag}(\boldsymbol{h}_{i}^{(t-1)''}  \odot (\boldsymbol{W}^{tT} \nabla_{\boldsymbol{b}^{t}} \xi_{i}))] \nonumber \\
        =& (\boldsymbol{W}^{tT} \mathbb{E}_{i}[ \nabla_{\boldsymbol{b}^{t}}^{2} \xi_{i}] \boldsymbol{W}^{t}) \odot \text{EhhT}^{(t-1)'} \nonumber \\
         & + \mathbb{E}_{i}[ \text{diag}(\boldsymbol{h}_{i}^{(t-1)''}  \odot (\boldsymbol{W}^{tT} \nabla_{\boldsymbol{b}^{t}} \xi_{i}))]. \label{eq:PropagateHessian_mathbb{E}}
\end{align}
where $\text{EhhT}^{(t-1)'} = \mathbb{E}_{i} [ \boldsymbol{h}_{i}^{(t-1)'} \otimes \boldsymbol{h}_{i}^{(t-1)'T} ] $.
The difference between the original and the approximate Hessian matrices in Eq.~(\ref{eq:PropagateHessian_mathbb{E}}) is bounded by
\begin{align}
      &   \left\| \text{Ele-Cov}( \boldsymbol{W}^{tT} \nabla_{\boldsymbol{b}^{t}}^{2} \xi_{i} \boldsymbol{W}^{t}, \boldsymbol{h}_{i}^{(t-1)'} \otimes \boldsymbol{h}_{i}^{(t-1)'T} ) \right\|_{F}^{2} \nonumber\\
  \leq&  L^{4} \sum_{\mu, \nu} \text{Var}([\boldsymbol{W}^{tT} \nabla_{\boldsymbol{b}^{t}}^{2} \xi_{i} \boldsymbol{W}^{t}]_{\mu \nu}), \label{eq:Hessian_Error_Bound_with_L_Lipschitz}
\end{align}
where $L$ is the Lipschitz constant of activation functions. 
For example, $L_{\text{ReLU}}$ and $L_{\text{sigmoid}}$ are $1$ and $0.25$, respectively. 
The details are in the Supplementary Materials D.

\section{Deriving the Newton Direction}
\label{sec:NNNewtonDirection}

Now we present a computationally feasible method to train FCNNs with Newton directions. 
First, we explain the ways to construct a PCH matrix.
Based on PCH matrices, we propose an efficient CG-based method incorporating the expectation approximation to derive Newton directions for multiple training instances, which we call EA-CG.
Finally, we provide an analysis of the space and time complexity for EA-CG. 

\subsection{PCH Matrix} 
\label{sec:BDA-PCH}

Based on our layer-wise equations in the previous section and the integration of expectation approximation, we can construct block matrices that vary in size and are located at the diagonal of the Hessian matrix.
We denote this block-diagonal matrix $\mathbb{E}_{i}[\overline{\nabla_{\boldsymbol{\theta}}^{2}}\xi_{i}]$ as $\text{diag}(\mathbb{E}_{i}[\nabla_{\boldsymbol{W}^{1}}^{2} \xi_{i}], \mathbb{E}_{i}[\nabla_{\boldsymbol{b}^{1}}^{2} \xi_{i}], \ldots, \mathbb{E}_{i}[\nabla_{\boldsymbol{W}^{k}}^{2} \xi_{i}], \mathbb{E}_{i}[\nabla_{\boldsymbol{b}^{k}}^{2} \xi_{i}])$.
Please note that $\mathbb{E}_{i}[\overline{\nabla_{\boldsymbol{\theta}}^{2}}\xi_{i}]$ is a block-diagonal Hessian matrix and not the complete Hessian matrix.
According to the explanation for the three possibilities of update directions in aforementioned section, $\mathbb{E}_{i}[\overline{\nabla_{\boldsymbol{\theta}}^{2}}\xi_{i}]$ is required to be modified.
Thus, we replace $\mathbb{E}_{i}[\overline{\nabla_{\boldsymbol{\theta}}^{2}}\xi_{i}]$ with $ \text{diag}(\mathbb{E}_{i}[\widehat{\nabla_{\boldsymbol{W}^{1}}^{2}} \xi_{i}], \mathbb{E}_{i}[\widehat{\nabla_{\boldsymbol{b}^{1}}^{2}} \xi_{i}], \ldots, \mathbb{E}_{i}[\widehat{\nabla_{\boldsymbol{W}^{k}}^{2}} \xi_{i}], \mathbb{E}_{i}[\widehat{\nabla_{\boldsymbol{b}^{k}}^{2}} \xi_{i}]) $ and denote the modified result as $\mathbb{E}_{i}[\widehat{\nabla_{\boldsymbol{\theta}}^{2}}\xi_{i}]$, where 
\begin{subequations}
\begin{align}
     \mathbb{E}_{i}[\widehat{\nabla_{\boldsymbol{b}^{k}}^{2}} \xi_{i}]  
  =& \text{Pos-Eig}( \mathbb{E}_{i}[\nabla_{\boldsymbol{h}_{i}^{k}}^{2} \xi_{i} ]) \label{eq:BDA-PCH-1} \\
     \mathbb{E}_{i}[\widehat{\nabla_{\boldsymbol{b}^{t-1}}^{2}} \xi_{i}]
  =& (\boldsymbol{W}^{tT} \mathbb{E}_{i}[ \widehat{\nabla_{\boldsymbol{b}^{t}}^{2}} \xi_{i}] \boldsymbol{W}^{t}) \odot \text{EhhT}^{(t-1)'}  \label{eq:BDA-PCH-2} \\
  +& \text{Pos-Eig}( \text{diag}(\mathbb{E}_{i}[\boldsymbol{h}_{i}^{(t-1)''} \odot (\boldsymbol{W}^{tT} \nabla_{\boldsymbol{b}^{t}} \xi_{i})]) ) \nonumber \\
     \mathbb{E}_{i}[\widehat{\nabla_{\boldsymbol{W}^{t}}^{2}} \xi_{i}] 
  =& \text{EhhT}^{t-1} \otimes \mathbb{E}_{i}[\widehat{\nabla_{\boldsymbol{b}^{t}}^{2}} \xi_{i}]. \label{eq:BDA-PCH-3}
\end{align}
\end{subequations}
We call $\mathbb{E}_{i}[\widehat{\nabla_{\boldsymbol{\theta}}^{2}}\xi_{i}]$ the block-diagonal approximation of positive curvature Hessian (PCH) matrix.
Any PCH matrix can be guaranteed to be PSD, which we explain in the following.

In order to show $\mathbb{E}_{i}[\widehat{\nabla_{\boldsymbol{\theta}}^{2}}\xi_{i}]$ is PSD, we have to show that for any $t$, both $\mathbb{E}_{i}[\widehat{\nabla_{\boldsymbol{b}^{t}}^{2}} \xi_{i}]$ and $\mathbb{E}_{i}[\widehat{\nabla_{\boldsymbol{W}^{t}}^{2}} \xi_{i}]$ are PSD. 
First, we consider the block matrix $\mathbb{E}_{i}[\nabla_{\boldsymbol{h}_{i}^{k}}^{2} \xi_{i}]$ that is a $n_{k}$ by $n_{k}$ square matrix in Eq.~(\ref{eq:BDA-PCH-1}).
If the criterion function $C(\hat{\boldsymbol{y}}_{i} \mid \boldsymbol{y}_{i})$ is convex, $\mathbb{E}_{i}[\nabla_{\boldsymbol{h}_{i}^{k}}^{2} \xi_{i}]$ is a PSD matrix. 
Otherwise, we decompose the matrix and replace the negative eigenvalues.
Fortunately $n_{k}$ is usually not very large, so $\mathbb{E}_{i}[\nabla_{\boldsymbol{h}_{i}^{k}}^{2} \xi_{i}]$ can be decomposed quickly\footnote{In our experience, the decomposition of an $1000 \times 1000$ matrix can be done within a few seconds in PyTorch.} and modified to a PSD matrix $\mathbb{E}_{i}[\widehat{\nabla_{\boldsymbol{h}_{i}^{k}}^{2}} \xi_{i}]$. 
Second, suppose that $\mathbb{E}_{i}[\widehat{\nabla_{\boldsymbol{b}^{t}}^{2}} \xi_{i}]$ is a PSD matrix, then $(\boldsymbol{W}^{tT} \mathbb{E}_{i}[ \widehat{\nabla_{\boldsymbol{b}^{t}}^{2}} \xi_{i}] \boldsymbol{W}^{t}) \odot \text{EhhT}^{(t-1)'}$ is PSD. 
Consequently, the negative eigenvalues of $\mathbb{E}_{i}[\widehat{\nabla_{\boldsymbol{b}^{t}}^{2}} \xi_{i}]$ stems form the diagonal part $\text{diag}(\mathbb{E}_{i}[\boldsymbol{h}_{i}^{(t-1)''} \odot (\boldsymbol{W}^{tT} \nabla_{\boldsymbol{b}^{t}} \xi_{i})])$, so we take the $\text{Pos-Eig}$ function for this diagonal part in Eq.~(\ref{eq:BDA-PCH-2}).
Third, because the Kronecker product of two PSD matrices is PSD, it implies $\mathbb{E}_{i}[\widehat{\nabla_{\boldsymbol{W}^{t}}^{2}} \xi_{i}]$ is PSD.

\subsection{Solving Linear Equation via EA-CG}
\label{sec:Single_Instance}

%Table for section 5
\begin{table*}[t]
  \caption{The comparisons of second-order methods.}
  \label{table:comparison}
  \centering
  \begin{tabular}{| l | l | l | l |}
    \hline
                    &    \cite{JM15a}               &      \cite{AB17a}         &      Ours         \\
    \hline
    Criterion Func. &     Non-Convex                 &         Convex             &    Non-Convex     \\
    \hline
    Curvature Info. &  Fisher                        & Gauss-Newton               &      PCH           \\
    Time Required   & $O(|\text{Batch}| \times \sum_{t=1}^{k} n_{t}^{2}) $ & $O(\sum_{t=2}^{k} n_{t-1}^{2} n_{t}) $ & $O(\sum_{t=2}^{k} n_{t-1}^{2} n_{t}) $ \\
    \hline
    Solving Scheme  &        KFI                     &          KFI               &    EA-CG \\
    Space Required  & $O(\sum_{t=0}^{k-1} n_{t}^{2} + \sum_{t=1}^{k} n_{t}^{2}) $ & $O(\sum_{t=0}^{k-1} n_{t}^{2} + \sum_{t=1}^{k} n_{t}^{2}) $ & $O( \sum_{t=0}^{k-1} n_{t} + \sum_{t=1}^{k} n_{t}^{2}) $ \\
    Time Required   & $O(\sum_{t=1}^{k}[ n_{t}^{3} + n_{t-1}^{3}+n_{t-1}^{2} n_{t} $ & $O(\sum_{t=1}^{k}[ n_{t}^{3} + n_{t-1}^{3}+n_{t-1}^{2} n_{t}$ & $O(|\text{CG}| \times \sum_{t=1}^{k}[n_{t}^{2}+ 2n_{t}n_{t-1} ])$ \\
                    & $\text{ }+ n_{t-1}n_{t}^{2}] ) $ & $\text{ } + n_{t-1}n_{t}^{2}] ) $ &  \\
    \hline
  \end{tabular}
\end{table*}

After obtaining a PCH matrix $\mathbb{E}_{i}[\widehat{\nabla_{\boldsymbol{\theta}}^{2}}\xi_{i}]$, we derive the update direction by solving the linear equation 
\begin{equation}
  \label{eq:Newton_Direction_Linear_Equation_Modified_with_diagonal}
  ( (1-\alpha) \mathbb{E}_{i}[\widehat{\nabla_{\boldsymbol{\theta}}^{2}}\xi_{i}] + \alpha I ) \boldsymbol{d}_{\boldsymbol{\theta}} = -\mathbb{E}_{i}[\nabla_{\boldsymbol{\theta}} \xi_{i}],
\end{equation}
where $0 < \alpha < 1$ and $\boldsymbol{d}_{\boldsymbol{\theta}} = [ \boldsymbol{d}_{\boldsymbol{W}^{1}}^{T} \text{ } \boldsymbol{d}_{\boldsymbol{b}^{1}}^{T} \cdots \boldsymbol{d}_{\boldsymbol{W}^{k}}^{T} \text{ } \boldsymbol{d}_{\boldsymbol{b}^{k}}^{T}]^{T}$.
Here, we use the weighted average of $\mathbb{E}_{i}[\widehat{\nabla_{\boldsymbol{\theta}}^{2}}\xi_{i}]$ and an identity matrix $I$ because this average turns the coefficient matrix of Eq.~(\ref{eq:Newton_Direction_Linear_Equation_Modified_with_diagonal}) that is PSD to positive definite and thus makes the solutions more stable. 
Due to the essence of diagonal blocks, Eq.~(\ref{eq:Newton_Direction_Linear_Equation_Modified_with_diagonal}) can be decomposed as
\begin{subequations}
\begin{align}
  ( (1 - \alpha) \mathbb{E}_{i}[\widehat{\nabla_{\boldsymbol{b}^{t}}^{2}} \xi_{i}] + \alpha I ) \boldsymbol{d}_{\boldsymbol{b}^{t}} = 
  & -\mathbb{E}_{i}[\nabla_{\boldsymbol{b}^{t}} \xi_{i}] \label{eq:LinearEquation-bias}\\
  ( (1 - \alpha) \mathbb{E}_{i}[\widehat{\nabla_{\boldsymbol{W}^{t}}^{2}} \xi_{i}] + \alpha I ) \boldsymbol{d}_{\boldsymbol{W}^{t}} = 
  & -\text{Vec}( \mathbb{E}_{i}[\nabla_{\boldsymbol{W}^{t}} \xi_{i}] ), \label{eq:LinearEquation-weight}
\end{align}
\end{subequations}
for $ t = 1, \ldots, k$.
To solve Eq.~(\ref{eq:LinearEquation-bias}), we attain the solutions by using the CG method directly.
For Eq.~(\ref{eq:LinearEquation-weight}), storing $\mathbb{E}_{i}[\widehat{\nabla_{\boldsymbol{W}^{t}}^{2}} \xi_{i}]$ is not an efficient way, so we apply the equation $(C^{T} \otimes A) \text{Vec}(B) = \text{Vec}(ABC)$ and Eq.~(\ref{eq:PropagateHessian_W_mathbb{E}}) to have the Hessian-vector product with a given vector $\text{Vec} (P)$:
\begin{subequations}
\begin{align}
   &        \mathbb{E}_{i}[\widehat{\nabla_{\boldsymbol{W}^{t}}^{2}} \xi_{i}] \text{Vec}(P) \nonumber \\
  =&       \text{Vec}( \mathbb{E}_{i}[\widehat{\nabla_{\boldsymbol{b}^{t}}^{2}} \xi_{i}] \cdot P \cdot 
           \mathbb{E}_{i}[ \boldsymbol{h}_{i}^{t-1} \otimes \boldsymbol{h}_{i}^{(t-1)T}] ) \label{eq:CG_approx-a} \\
  \approx& \text{Vec}( \mathbb{E}_{i}[\widehat{\nabla_{\boldsymbol{b}^{t}}^{2}} \xi_{i}] \cdot P \cdot 
           \mathbb{E}_{i}[ \boldsymbol{h}_{i}^{t-1} ] \otimes \mathbb{E}_{i}[\boldsymbol{h}_{i}^{(t-1)T}] ). \label{eq:CG_approx-b}
\end{align}
\end{subequations}

Based on Eq.~(\ref{eq:CG_approx-a}), we derive the Hessian-vector products of $\mathbb{E}_{i}[\widehat{\nabla_{\boldsymbol{W}^{t}}^{2}} \xi_{i}]$ via $\mathbb{E}_{i}[\widehat{\nabla_{\boldsymbol{b}^{t}}^{2}} \xi_{i}]$, thereby reducing the space complexity of storing the curvature information.
Furthermore, applying the CG method becomes much more efficient with Eq.~(\ref{eq:CG_approx-b}), which we call EA-CG.
The details are elaborated in the following subsections.

\subsection{Analysis of Space Complexity}
\label{sec:AnalysisSpaceComplexity}

By considering the devised method in aforementioned subsection, we analyze the required space complexity to store the distinct types of the curvature information.
The original Newton's method requires space to store $\mathbb{E}_{i}[\nabla_{\boldsymbol{\theta}}^{2} \xi_{i}]$, so that its space complexity is $O( [ \sum_{t=1}^{k} n_{t-1}(n_{t}+1) ]^{2} )$. If we consider the PCH matrix $\mathbb{E}_{i}[\widehat{\nabla_{\boldsymbol{\theta}}^{2}} \xi_{i}]$, the space complexity turns out to be $O(  \sum_{t=1}^{k} [(n_{t-1}n_{t})^{2} + n_{t}^{2}]  )$. 
According to Eq.~(\ref{eq:CG_approx-a}), it is not necessary to store $\mathbb{E}_{i}[\widehat{\nabla_{\boldsymbol{W}^{t}}^{2}} \xi_{i}]$ anymore because $\mathbb{E}_{i}[\widehat{\nabla_{\boldsymbol{b}^{t}}^{2}} \xi_{i}]$ is sufficient to derive the solution to Eq.~(\ref{eq:Newton_Direction_Linear_Equation_Modified_with_diagonal}) with the CG method.
Thus, the space complexity is reduced to $O( \sum_{t=1}^{k}  n_{t}^{2} )$. 

\subsection{Analysis of Additional Time Complexity}
\label{sec:AnalysisTimeComplexity}

In this subsection, we elaborate on the additional time complexity introduced in our proposed method.
In contrast to the SGD, our method contributes more computation to the training process of FCNNs.
The extra computation mainly originates from two portions of our method: the propagation of the curvature information and the computation of EA-CG method.
To propagate the curvature information, we are required to perform the matrix-to-matrix products twice, which is more computationally expensive than propagating gradients.
The time complexity of propagating the curvature information with Eq.~(\ref{eq:PropagateHessian_mathbb{E}}) can be estimated as $O( \sum_{t=1}^{k} [n_{t-1} n_{t} (n_{t-1} + n_{t})] )$.
We also have the extra cost involved in applying EA-CG method.
The complexity of EA-CG method mainly stems from the Hessian-vector products Eq.~(\ref{eq:CG_approx-b}). 
Regarding Eq.~(\ref{eq:CG_approx-b}), we must conduct the matrix-vector products twice and the vector-vector outer products once in order to acquire the Hessian-vector product.
Thus, the time complexity pertaining to the truncated-Newton method is $O( |\text{CG}| \times \sum_{t=1}^{k} [ n_{t} (2 n_{t-1} + n_{t}) ] )$, where $|\text{CG}|$ is the number of iterations of EA-CG method.

\section{Related Works}
\label{sec:Related_Works}

%figure for section 6
\begin{figure*}[t]
    \centering
    \includegraphics[width=0.9\columnwidth]{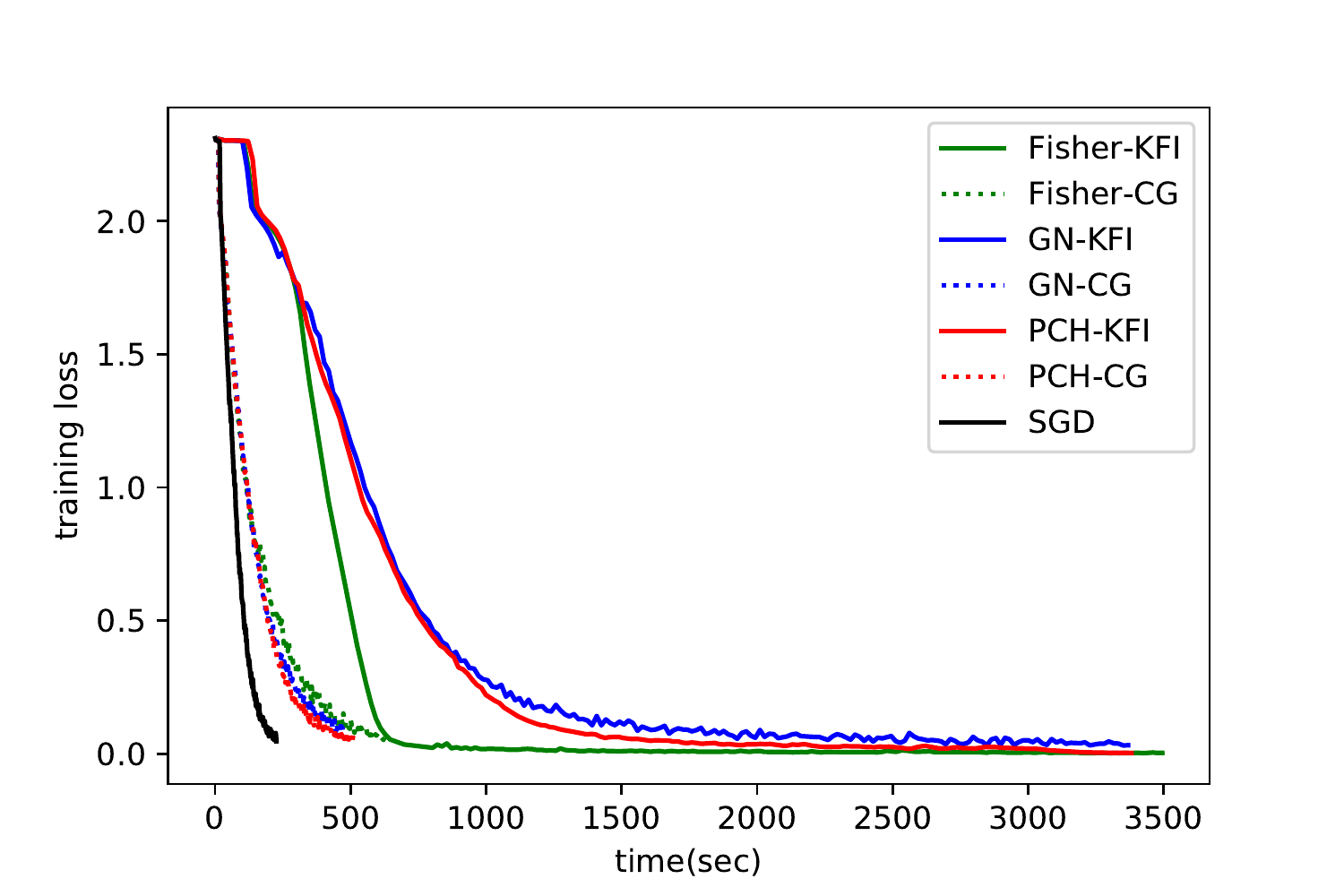}
    \includegraphics[width=0.9\columnwidth]{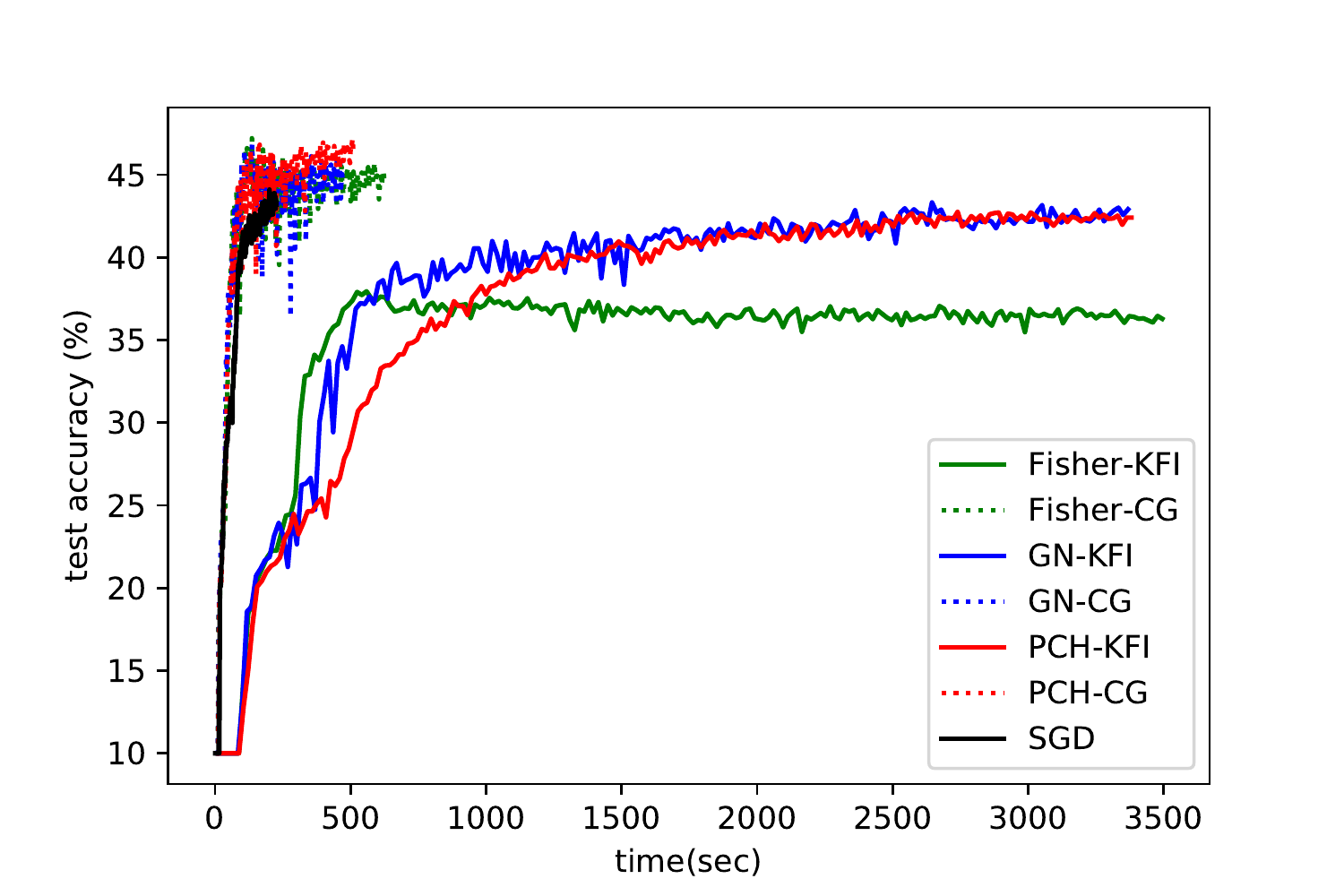}
  \caption{Comparison of different curvature information and solving methods for the convex criterion function ``cross-entropy'' on ``Cifar-10''.}
  \label{fig:Convex_cifar_time}
    \includegraphics[width=0.9\columnwidth]{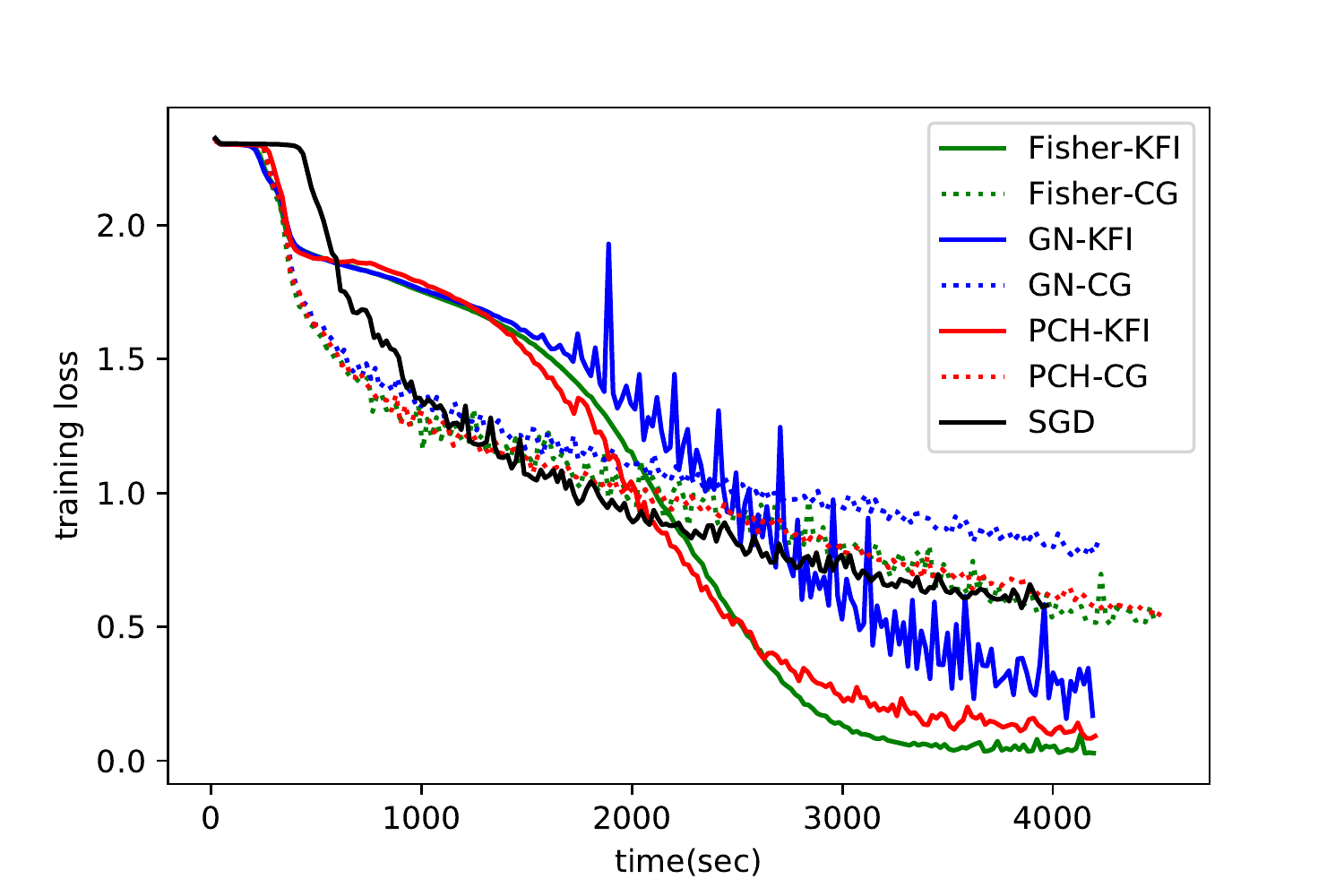}
    \includegraphics[width=0.9\columnwidth]{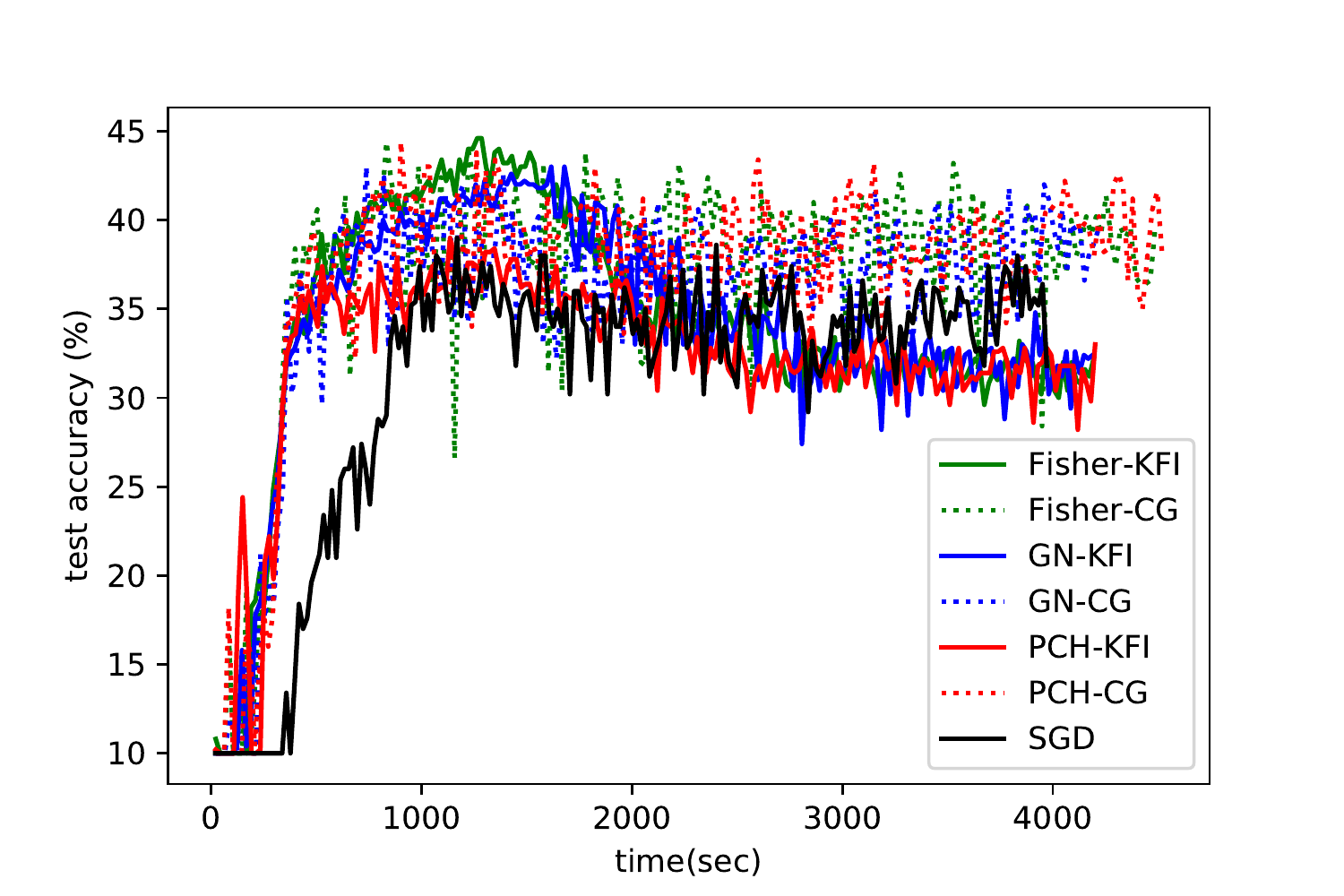}
  \caption{Comparison of different curvature information and solving methods for the convex criterion function ``cross-entropy'' on ``ImageNet-10''.}
  \label{fig:Convex_imagenet_time}
\end{figure*}

In this section, we elaborate on the differences between our work and two closely related works using the notation established in the previous sections.

\citeauthor{JM15a} developed KFAC by considering the FCNNs with the convex criterion and non-convex activation functions. 
KFAC utilized ($\mathbb{E}_{i} [\tilde{F}_{i}] + \alpha I$) where $\tilde{F}_{i}$ is the Fisher matrix (for any training instance $\boldsymbol{x}_{i}$) to measure the curvature and used the Khatri-Rao product to rewrite $\tilde{F}_{i}$, which yields the following equation:
\[
  [\tilde{F}_{i}]_{\mu \nu} = (\boldsymbol{h}_{i}^{\mu-1} \otimes \boldsymbol{h}_{i}^{(\mu-1)T}) \otimes ( \nabla_{\boldsymbol{b}^{\nu}} \xi_{i} \otimes \nabla_{\boldsymbol{b}^{\nu}} \xi_{i}^{T} ).
\]
Since it is difficult to find the inverse of $(\mathbb{E}_{i}[\tilde{F}_{i}] + \alpha I)$, KFAC substitutes a block-diagonal matrix $\hat{F}_{i}$ for $\tilde{F}_{i}$ and thus has the formulation:
\begin{equation*}
\begin{split}
          \mathbb{E}_{i}[\hat{F}_{i}^{t}] 
        =& \mathbb{E}_{i}[(\boldsymbol{h}_{i}^{t-1} \otimes \boldsymbol{h}_{i}^{(t-1)T}) \otimes ( \nabla_{\boldsymbol{b}^{t}} \xi_{i} \otimes \nabla_{\boldsymbol{b}^{t}} \xi_{i}^{T} )] \\
  \approx& \mathbb{E}_{i}[\boldsymbol{h}_{i}^{t-1} \otimes \boldsymbol{h}_{i}^{(t-1)T}] \otimes \mathbb{E}_{i}[\nabla_{\boldsymbol{b}^{t}} \xi_{i} \otimes \nabla_{\boldsymbol{b}^{t}} \xi_{i}^{T}],\\
\end{split}
\end{equation*}
where $t = 1, \ldots, k$.
To derive $(\mathbb{E}_{i}[\hat{F}_{i}^{t}] + \alpha I)^{-1}$ efficiently, KFAC comes up with the approximation
\begin{equation}
\label{eq:Fisher_Approx}
           (\mathbb{E}_{i}[\hat{F}_{i}^{t}] + \alpha I)^{-1} \approx  (\mathcal{H}^{t})^{-1} \otimes (\mathcal{G}^{t})^{-1}, 
\end{equation}
where $(\mathcal{H}^{t})^{-1} = (\mathbb{E}_{i}[\boldsymbol{h}_{i}^{t-1} \otimes \boldsymbol{h}_{i}^{(t-1)T}] + \pi_{t} \sqrt{\alpha} I)^{-1}$, $(\mathcal{G}^{t})^{-1} = (\mathbb{E}_{i}[\nabla_{\boldsymbol{b}^{t}} \xi_{i} \otimes \nabla_{\boldsymbol{b}^{t}} \xi_{i}^{T}] + (\sqrt{\alpha}/\pi_{t})  I )^{-1}$.
Therefore, the update directions of KFAC can be acquired via the following equation:
\begin{equation*}
\begin{split}
    \hat{\boldsymbol{d}}_{\boldsymbol{W}^{t}}
  =& - ((\mathcal{H}^{t})^{-1} \otimes (\mathcal{G}^{t})^{-1}) \cdot \text{Vec}(\mathbb{E}_{i}[\nabla_{\boldsymbol{W}^{t}} \xi_{i}])\\
  =& -\text{Vec}( (\mathcal{G}^{t})^{-1} \mathbb{E}_{i}[\nabla_{\boldsymbol{W}^{t}} \xi_{i}] (\mathcal{H}^{t})^{-1} ),\\
\end{split}
\end{equation*}
and we refer this type of inverse methods as Kronecker-Factored Inverse (KFI) methods in this paper.
In contrast, we derive the update directions by applying EA-CG method.
Moreover, KFAC uses the Fisher matrix, while we use the PCH matrix.

Based on the discussion of the FCNNs with convex criterion functions and piecewise linear activation functions, \citeauthor{AB17a} developed KFRA.
As a result, the diagonal term disappears in Eq.~(\ref{eq:PropagateHessian}).
Thus, the Gauss-Newton matrix becomes no different from the Hessian matrix.
KFRA also employs KFI method, and hence the update direction of KFRA can be derived from 
\[
  \tilde{\boldsymbol{d}}_{\boldsymbol{W}^{t}} = - \text{Vec}((\tilde{\mathcal{G}}^{t})^{-1} \mathbb{E}_{i}[\nabla_{\boldsymbol{W}^{t}} \xi_{i}] (\mathcal{H}^{t})^{-1}),
\]
where $(\tilde{\mathcal{G}}^{t})^{-1} = (\mathbb{E}_{i}[ \text{GN}(\nabla_{\boldsymbol{b}^{t}}^{2} \xi_{i} )] + (\sqrt{\alpha}/\pi_{t})  I )^{-1}$, and GN stands for Gauss-Newton matrix.
The method of deriving the update directions is again different between KFRA and our work.
In addition, KFRA still works under the non-convex activation functions, but it exists the difference between Gauss-Newton matrix and Hessian matrix.

Table~\ref{table:comparison} highlights the three main components of these two related works and our method.
As shown in Table~\ref{table:comparison}, KFRA cannot handle non-convex criterion functions due to the curvature information they utilized.
The Gauss-Newton matrix becomes indefinite if the criterion function is non-convex.
Note that the difference between the original and approximated matrices always exists in Eq.~(\ref{eq:Fisher_Approx}) if $\mathbb{E}_{i}[\hat{F}_{i}^{t}]$ is not diagonal.
Moreover, the time complexity of KFI method is similar to EA-CG method.

\section{Experimental Evaluation}
\label{sec:Experiments}

%figure for section 6
\begin{figure*}[t]
    \centering
    \includegraphics[width=0.9\columnwidth]{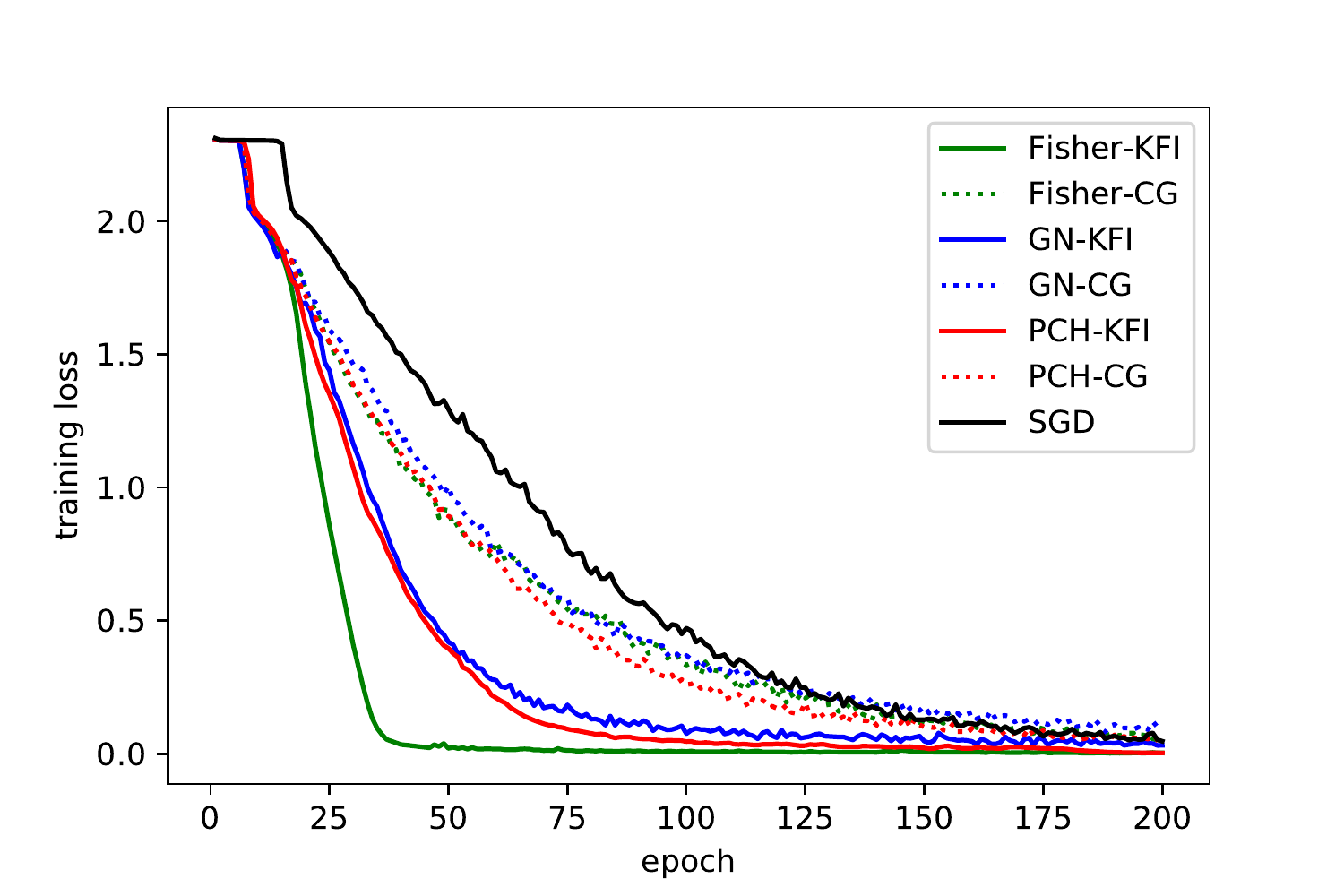}
    \includegraphics[width=0.9\columnwidth]{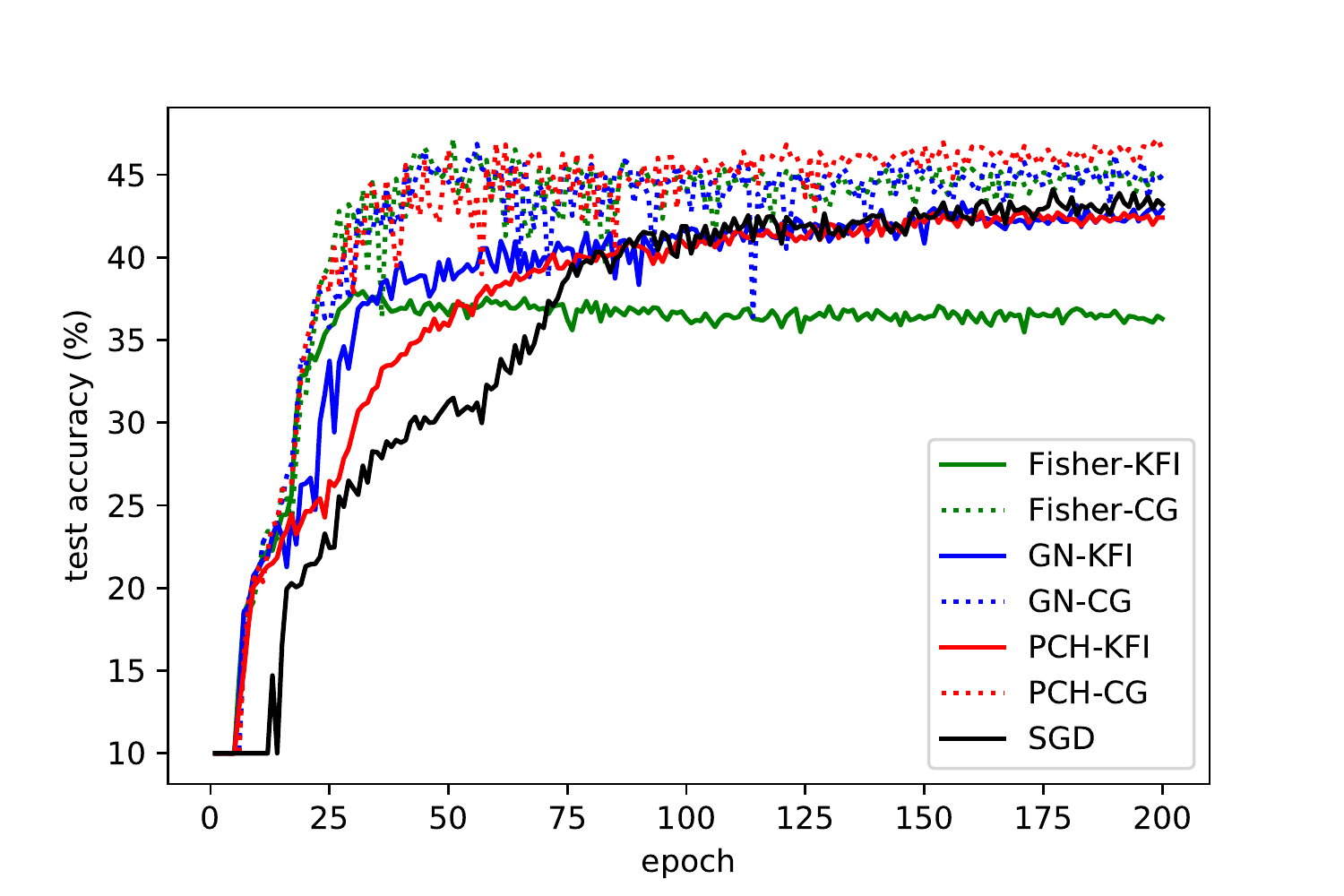}
  \caption{Comparison of different curvature information and solving methods for the convex criterion function ``cross-entropy'' on ``Cifar-10''.}
  \label{fig:Convex_cifar_epoch}
    \includegraphics[width=0.9\columnwidth]{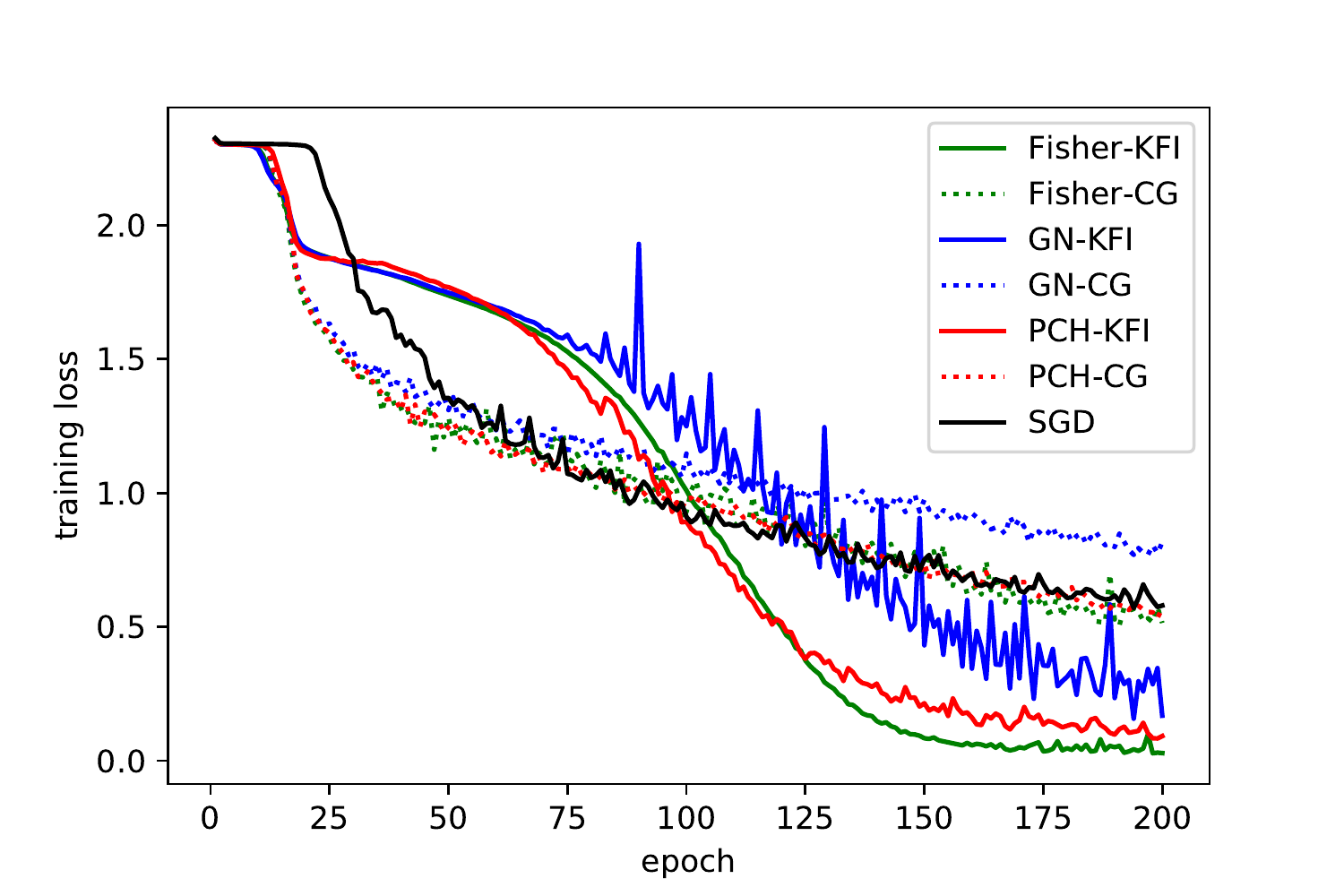}
    \includegraphics[width=0.9\columnwidth]{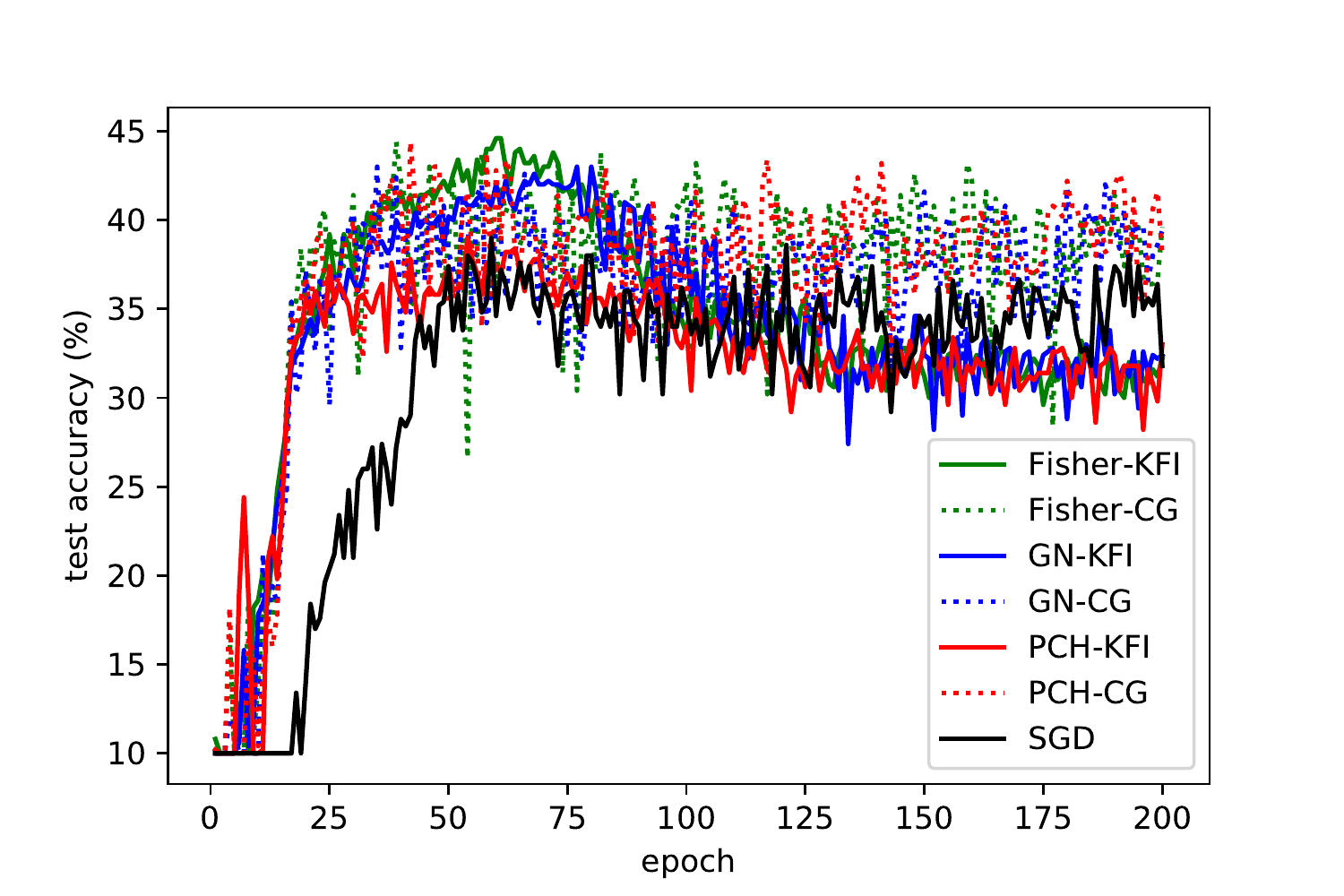}
  \caption{Comparison of different curvature information and solving methods for the convex criterion function ``cross-entropy'' on ``ImageNet-10''.}
  \label{fig:Convex_imagenet_epoch}
\end{figure*}

%figure for section 6
\begin{figure*}[t]
  \centering
    \includegraphics[width=0.9\columnwidth]{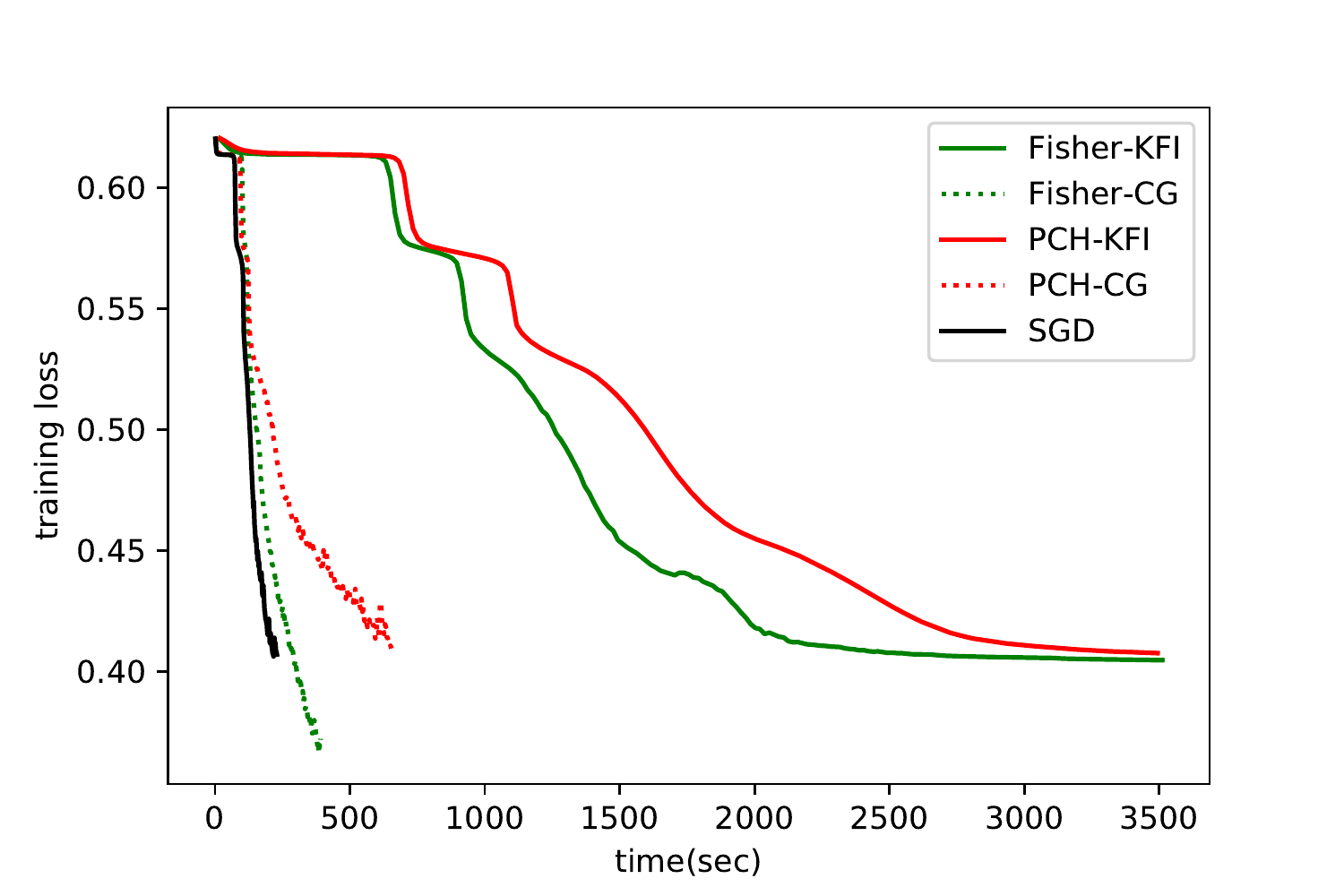}
    \includegraphics[width=0.9\columnwidth]{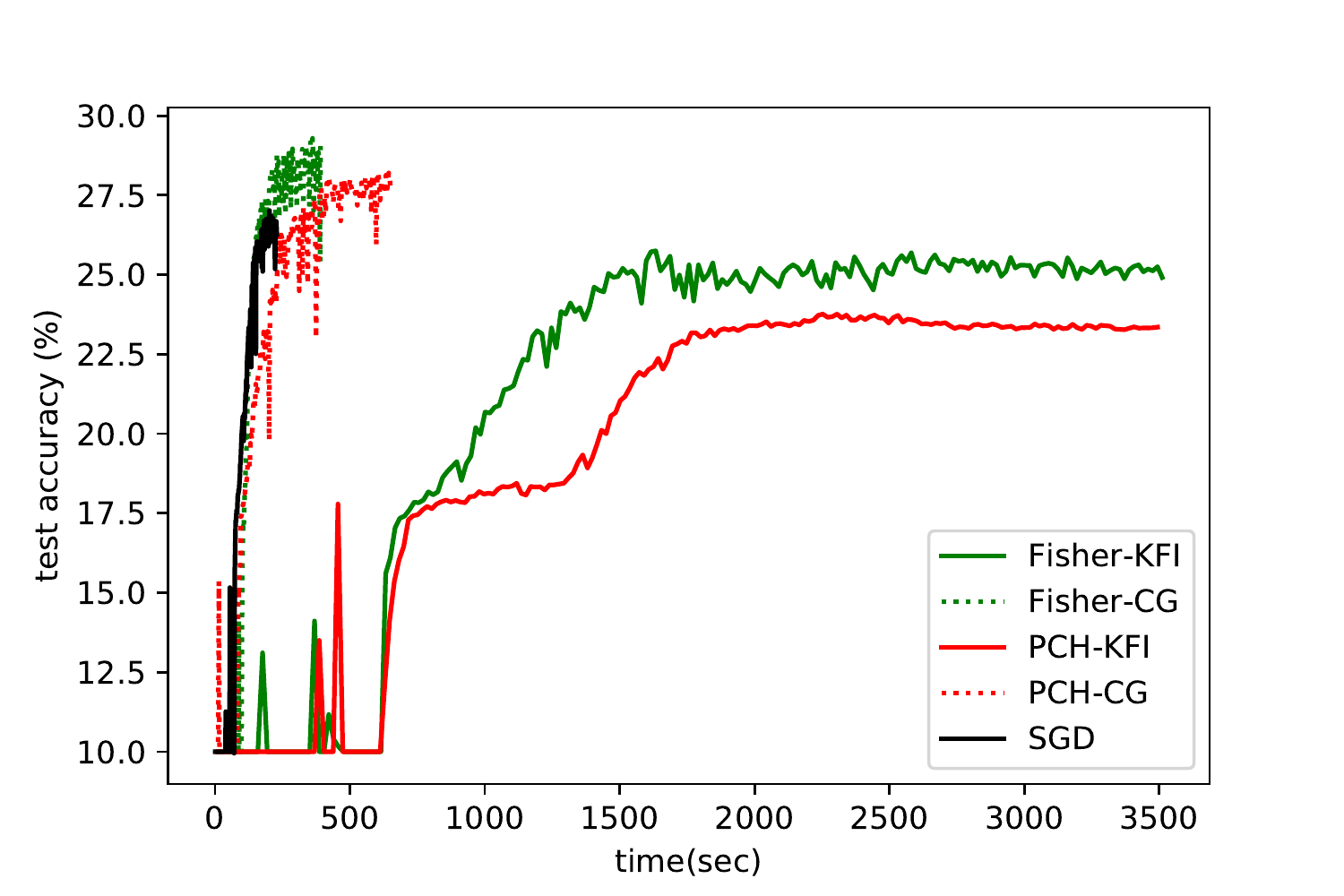}
  \caption{Comparison of different curvature information and solving methods for the non-convex criterion function ``Eq.~(\ref{eq:Non-Convex_Criterion})'' on ``Cifar-10''. The Gauss-Newton matrix is removed from this figure since it is not PSD in this type of experiments.}
  \label{fig:Nonconvex_cifar_time}
    \includegraphics[width=0.9\columnwidth]{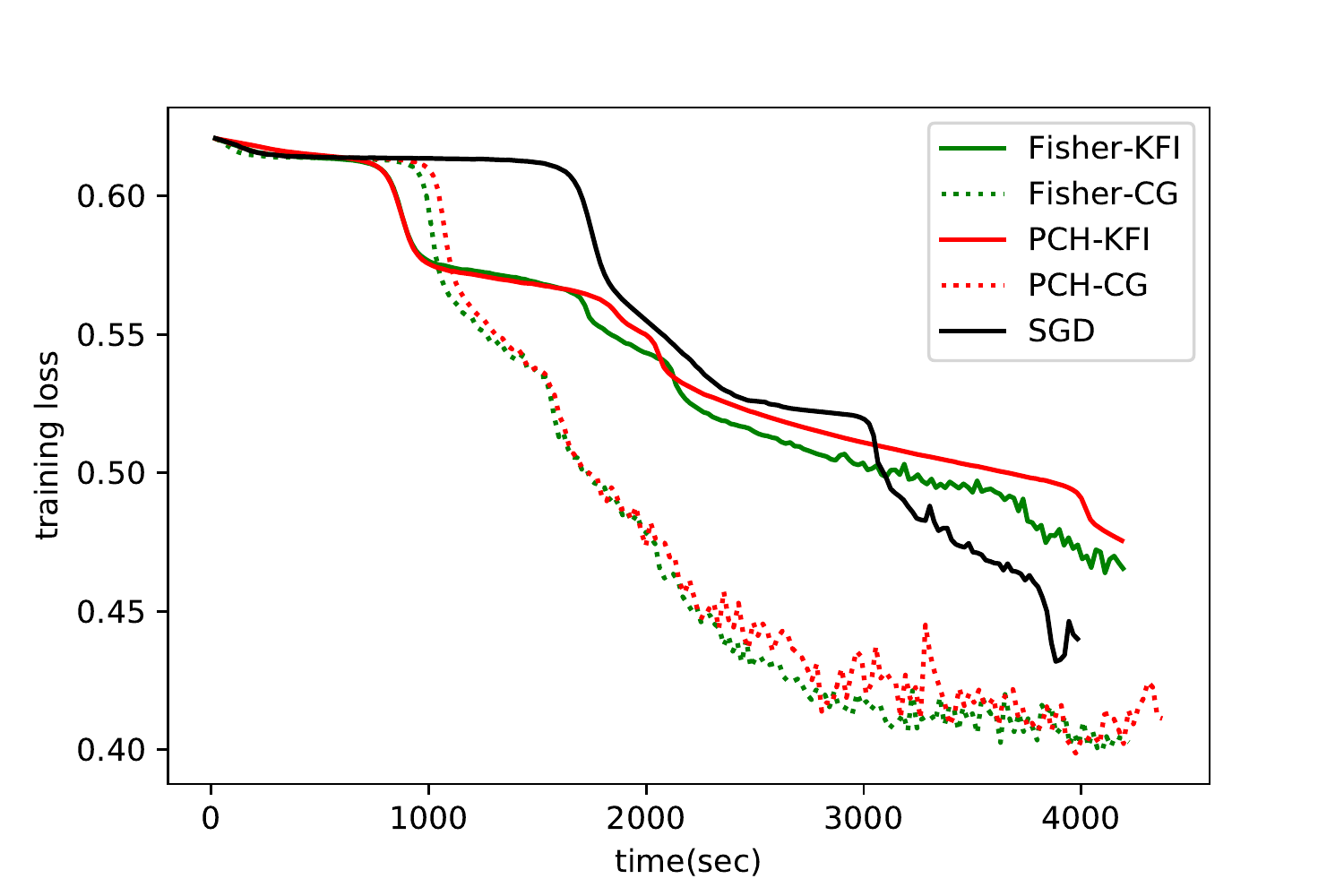}
    \includegraphics[width=0.9\columnwidth]{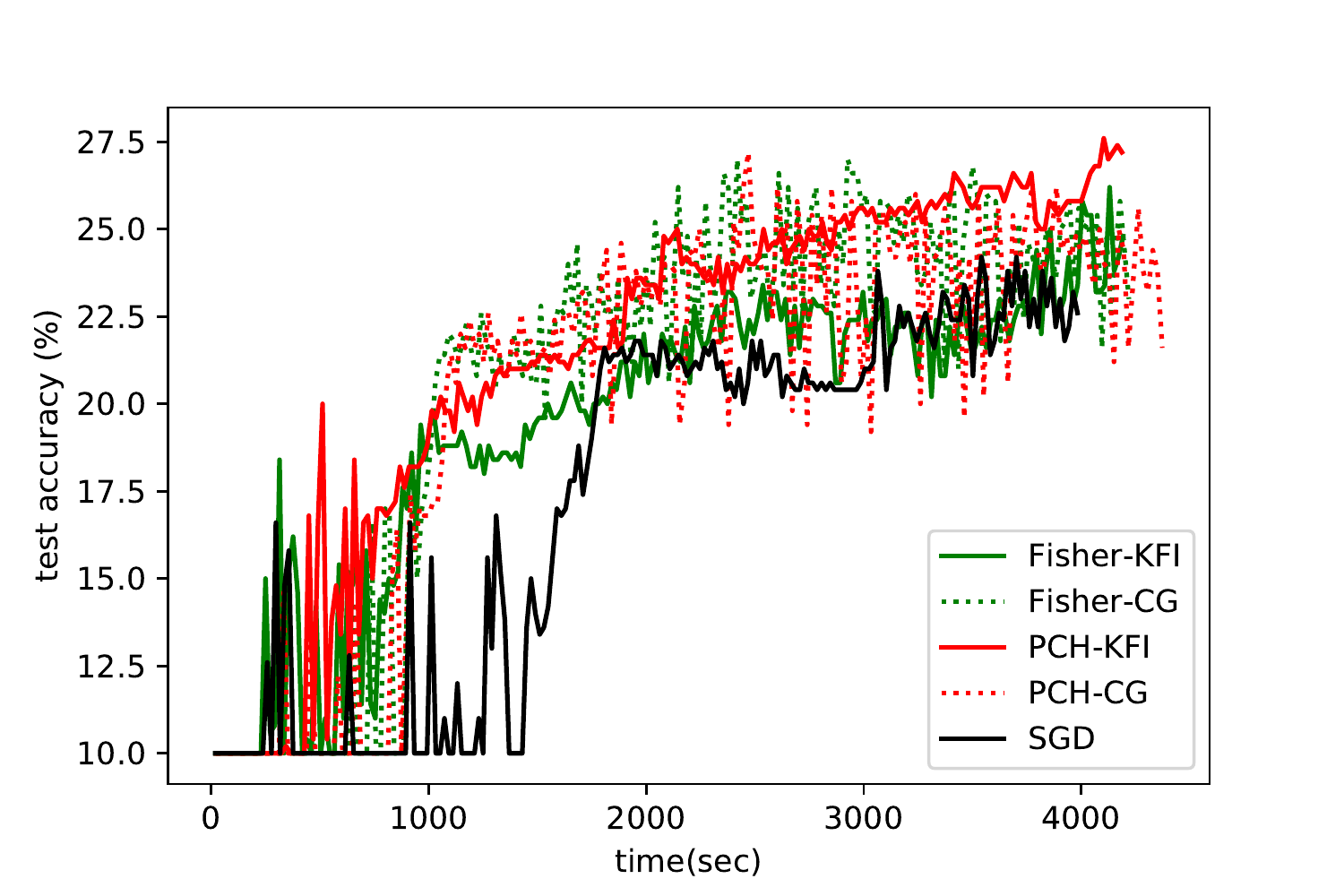}
  \caption{Comparison of different curvature information and solving methods for the non-convex criterion function ``Eq.~(\ref{eq:Non-Convex_Criterion})'' on ``ImageNet-10''. The Gauss-Newton matrix is removed from this figure since it is not PSD in this type of experiments.}
  \label{fig:Nonconvex_imagenet_time}
\end{figure*}

%figure for section 6
\begin{figure*}[t]
  \centering
    \includegraphics[width=0.9\columnwidth]{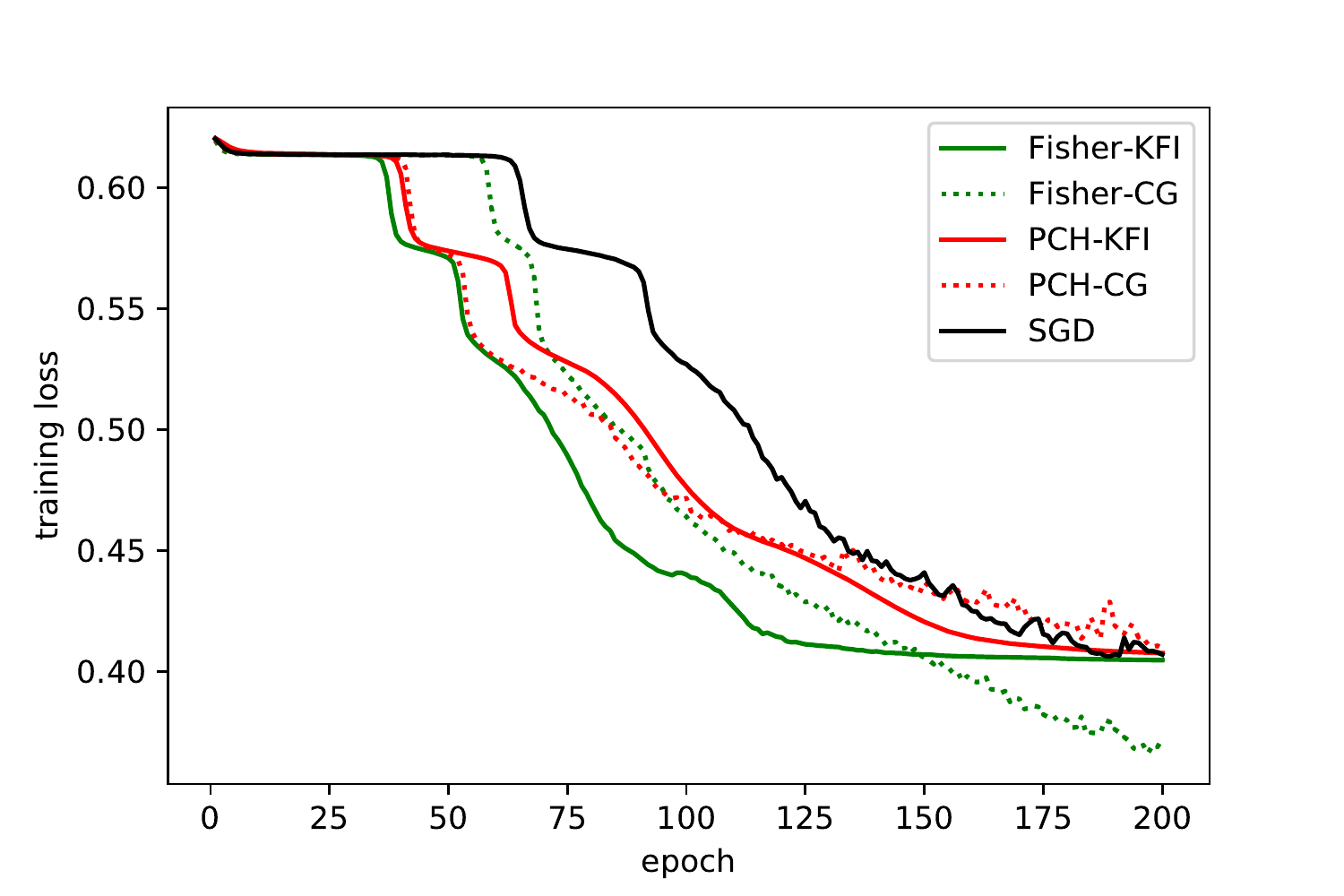}
    \includegraphics[width=0.9\columnwidth]{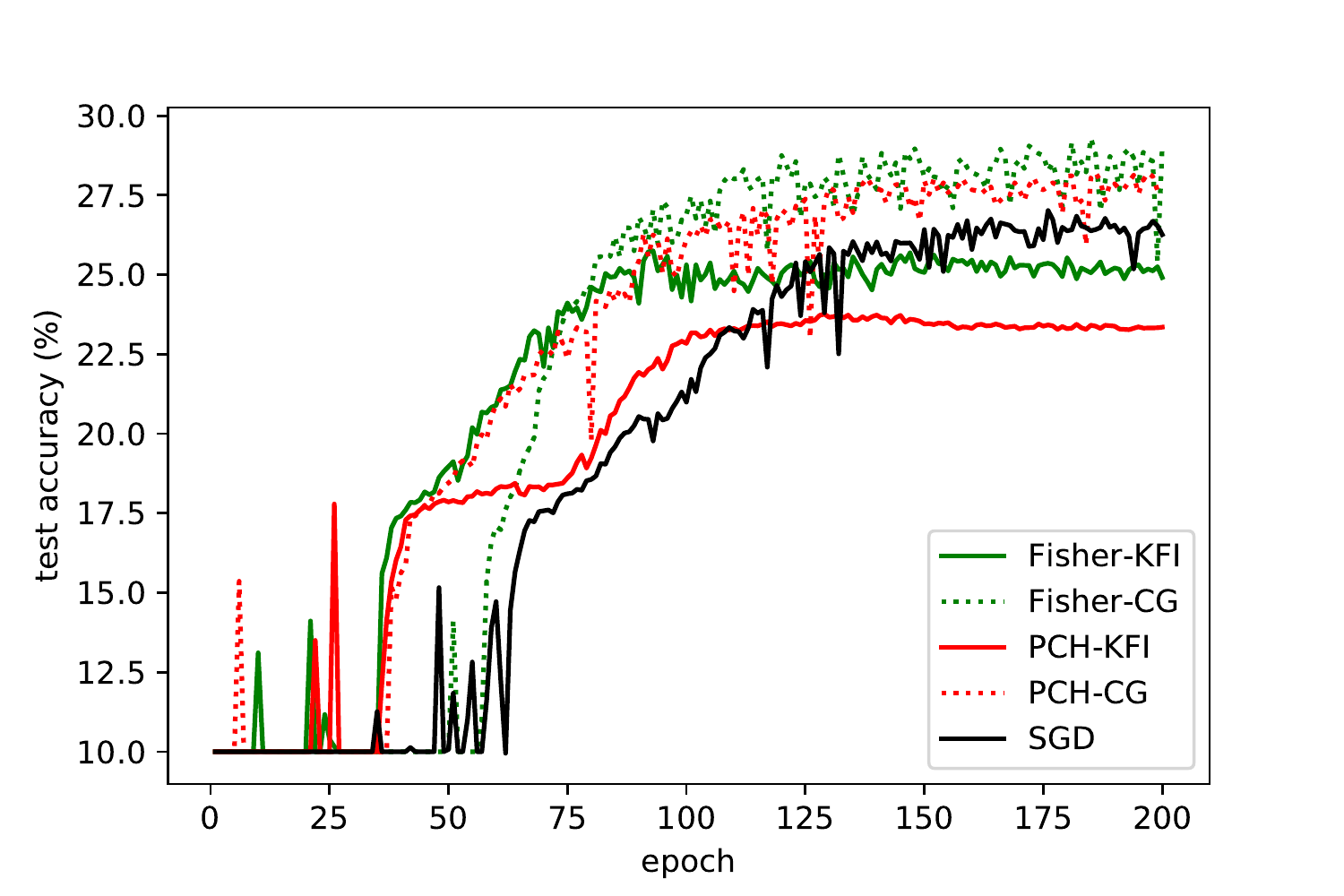}
  \caption{Comparison of different curvature information and solving methods for the non-convex criterion function ``Eq.~(\ref{eq:Non-Convex_Criterion})'' on ``Cifar-10''. The Gauss-Newton matrix is removed from this figure since it is not PSD in this type of experiments.}
  \label{fig:Nonconvex_cifar_epoch}
    \includegraphics[width=0.9\columnwidth]{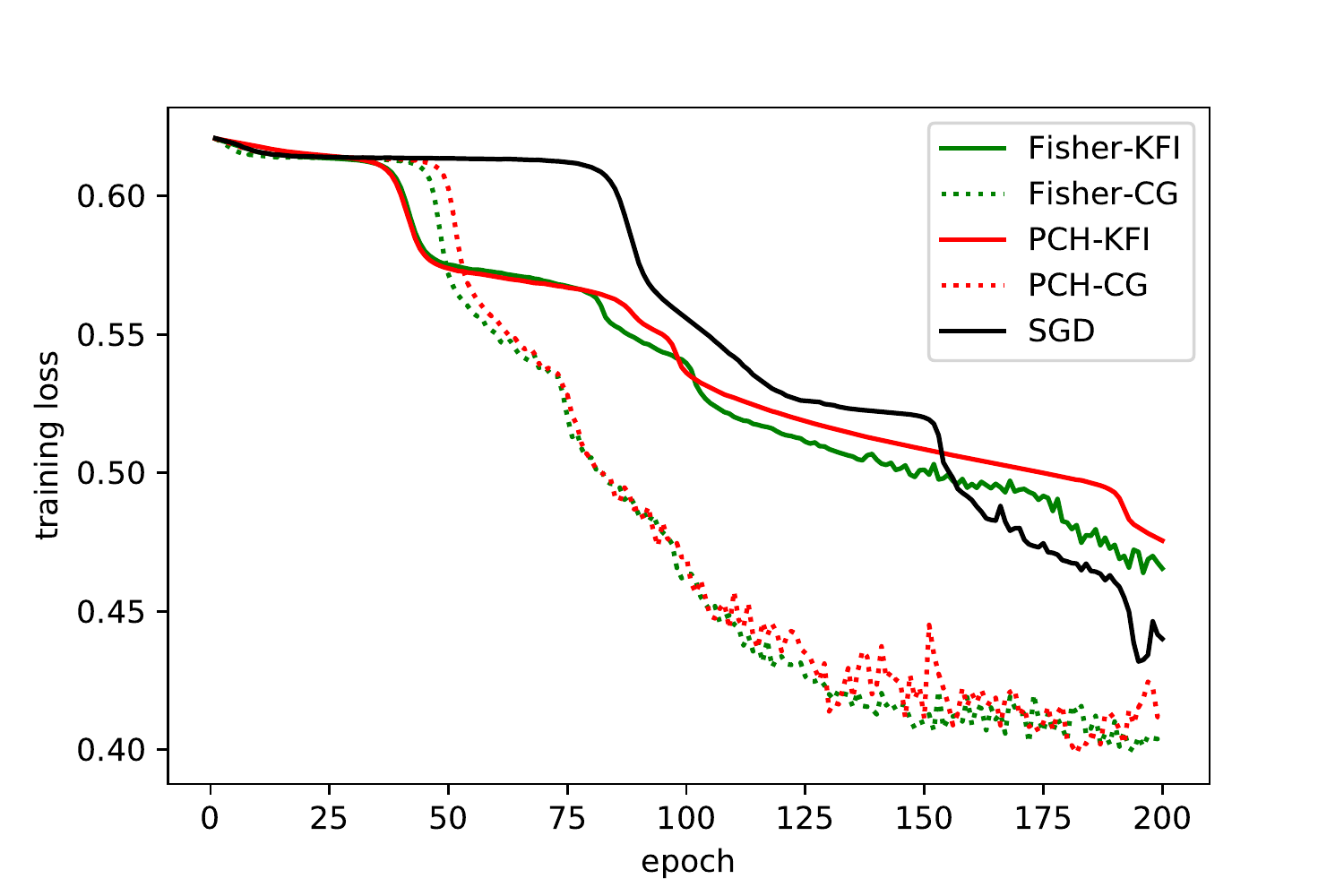}
    \includegraphics[width=0.9\columnwidth]{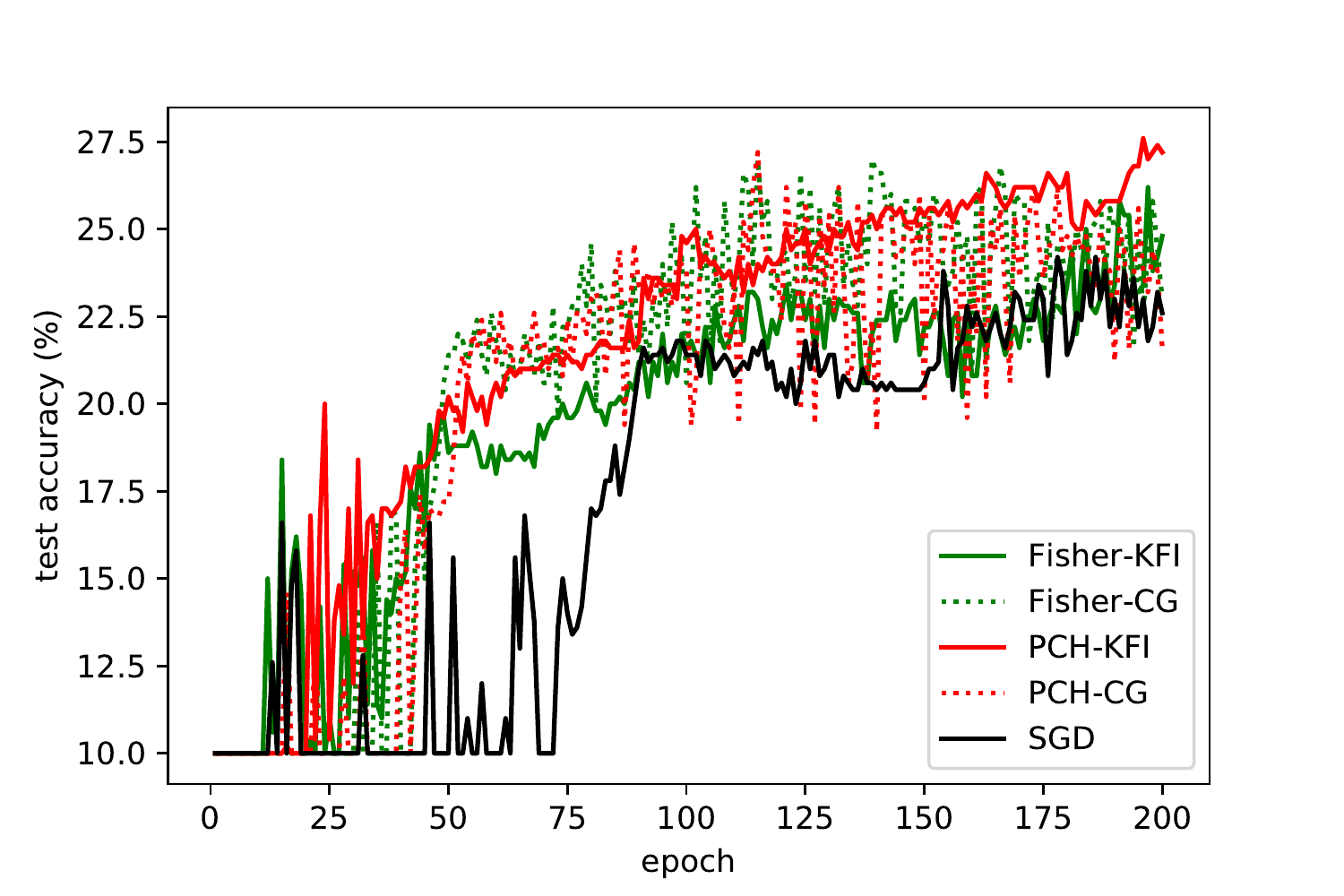}
  \caption{Comparison of different curvature information and solving methods for the non-convex criterion function ``Eq.~(\ref{eq:Non-Convex_Criterion})'' on ``ImageNet-10''. The Gauss-Newton matrix is removed from this figure since it is not PSD in this type of experiments.}
  \label{fig:Nonconvex_imagenet_epoch}
\end{figure*}

Our empirical studies aim to examine not only three different types of curvature information including the Fisher, Gauss-Newton and PCH matrices, but also the two solving methods, i.e., KFI and EA-CG methods, in terms of training loss and testing accuracy.
We encompassed SGD with momentum as the baseline and fixed its momentum to $0.9$.
For better comparison, we considered FCNNs with either convex or non-convex criterion functions and conducted our experiments\footnote{Experiments were implemented by using PyTorch libraries and run on a GTX-1080Ti GPU.} on the image datasets that comprise Cifar-10 \cite{AK09a} and ImageNet-10\footnote{We randomly choose ten classes from the ImageNet \cite{JD09a} dataset.}.
The network structures are ``3072-1024-512-256-128-64-32-16-10" and ``150528-1024-512-256-128-64-32-16-10" for Cifar-10 and ImageNet-10, respectively.
For the PCH matrices, we explored two possible scenarios of Eq.(\ref{eq:BDA-PCH-2}): 1) taking the absolute values of the diagonal part, i.e., $\gamma = -1$ and 2) applying $\max (x, 0)$ function to the diagonal part, i.e., $\gamma = 0$.
The first and second scenarios of the PCH matrices are separately dubbed PCH-1 and PCH-2.
Since PCH-1 and PCH-2 produced the similar results of training loss and testing accuracy\footnote{Detailed results are in Supplementary Material A}, we only reported PCH-1 in the corresponding figures below for the sake of simplicity. 
In all of our experiments, we used sigmoid as the non-convex activation function of FCNNs and trained the networks with 200 epochs, i.e., seeing the entire training samples 200 times.
Furthermore, we utilized ``Xavier" initialization method \cite{XG10a} and performed the grid search on $\text{learning rate} = [0.05, 0.1, 0.2]$, $|\text{Batch}| = [100, 500, 1000]$ and $\alpha = [0.01, 0.02, 0.05, 0.1]$. 
Regarding EA-CG method, we have to determine two hyper-parameters that control its stopping conditions.
One is the maximal iteration number, \text{max}$|\text{CG}|$; the other is the constant of the related error bound, $\epsilon_{\text{CG}}$. 
Thus, we also performed the grid search on $\text{max}|\text{CG}| = [5, 10, 20, 50]$ and $\epsilon_{\text{CG}} = [10^{-10}, 10^{-5}, 10^{-2}, 10^{-1}]$. 
Finally, to further investigate the different types of curvature information, we calculate the errors caused by approximating the true Hessian in a layer-wise fashion.

\subsection{Convex Criterion}
\label{sec:Experiments_Convex_Criterion}

In the first type of experiments, we examined the performance of different second-order methods.
According to the prior section, the differences between these second-order methods originate from two parts.
The first part is the curvature information, e.g., the Fisher matrix $\hat{F}_{i}$ in KFAC, and the other is the solving method for the linear equations.
In these experiments, we used cross-entropy as the convex criterion function.

It is noteworthy that the original KFI method runs out of GPU memory for ImageNet-10 because the dimension of the features ($n_{0}$) in ImageNet-10 is too high, which conforms with our analysis in Table~\ref{table:comparison}. Thus, we utilized
\begin{equation}
\label{eq:KFI-approx}
  \mathcal{H}^{1} \approx \mathbb{E}_{i}[\boldsymbol{h}_{i}^{0}] \otimes \mathbb{E}_{i}[\boldsymbol{h}_{i}^{0T}] + \pi_{1} \sqrt{\alpha} I
\end{equation}
and Sherman-Morrison formula to derive the update directions. 
Albeit this approximation worked in our ImageNet-10 experiments, we observed that this approximation is not stable for other datasets and models that we have explored. 

\subsubsection{Wall Clock Time}
As shown in Figure~\ref{fig:Convex_cifar_time}, EA-CG method converges faster with respect to wall clock time and has better performance of testing accuracy than KFI method for Cifar-10. 
The runtime behavior adheres to Table~\ref{table:comparison}.
We also notice that the different types of curvature information with EA-CG method have the similar performance.
In contrast, KFI method using the Fisher matrix converges faster than using the other two types of matrices.
For ImageNet-10, Figure~\ref{fig:Convex_imagenet_time} exhibits that EA-CG and KFI methods take almost the same amount of time for $200$ epochs. 
Besides, as shown in Figure~\ref{fig:Convex_imagenet_time}, KFI method has lower training loss but may suffer from the overfitting issue.
Please note that we applied Eq~(\ref{eq:KFI-approx}) to the first layer of the neural network for KFI method. 
Otherwise, the original KFI method ran out of GPU memory for ImageNet-10.
However, applying this approximation reduces the memory usage of KFI method but expedites KFI method accordingly.

It is worth noting that second-order methods performed comparable to SGD in Figure~\ref{fig:Convex_imagenet_time}, but the phenomenon is completely different in Figure~\ref{fig:Convex_cifar_time}.
This is because the image sizes of Cifar-10 and ImageNet-10 are varied. 
The image size of ImageNet-10 is $224 \times 224 \times 3$, and hence propagating the gradients of the weights backward is much more expensive than it in Cifar-10 whose image size is $32 \times 32 \times 3$. 
Based on this premise, the extra cost of propagating our proposed PCH where only the Hessian of the bias terms is required to propagate is comparatively low.
Similarly, the cost of solving the update directions is comparatively low.

\subsubsection{Epoch}
As shown in Figure~\ref{fig:Convex_cifar_epoch}, KFI method provides a precise descent direction of each epoch, so the training loss decreases faster in terms of epochs.
In contrast, Figure~\ref{fig:Convex_cifar_time} shows KFI method converges slower in terms of wall clock time, which varied with different implementation.
We argue that we have already done our best effort to implement KFI method by using the inverse function in PyTorch.
As shown in either Figure~\ref{fig:Convex_cifar_time} or Figure~\ref{fig:Convex_cifar_epoch}, EA-CG method has the better results of testing accuracy for Cifar-10.
For ImageNet-10, the results of Figure~\ref{fig:Convex_imagenet_epoch} are consistent with the results of Figure~\ref{fig:Convex_imagenet_time}.
Thus, we do not repeat the same narrative here. 

\subsection{Non-Convex Criterion}

In order to demonstrate our capability of handling non-convex criterion functions, we designed the second type of experiments.
In these experiments, we considered the following criterion function:
\begin{equation}
\label{eq:Non-Convex_Criterion} 
  C(\hat{\boldsymbol{y}_{i}} \mid \boldsymbol{y}_{i}) = \frac{1}{1 + e^{\delta ( \boldsymbol{y}_{i}^{T}\hat{\boldsymbol{y}_{i}} - \epsilon)} },
\end{equation}
where we fix $\delta = 5$ and $\epsilon = 0.2$.
This criterion function implies that the upper bound of the loss function exists. 
Apart from the criterion function, we followed the same settings as those used in the first type of experiments.
Note that the Gauss-Newton matrix is not PSD if the criterion function of training FCNNs is non-convex.
Thus, we excluded the Gauss-Newton matrix from this comparison and focused on comparing the other two types of matrices with either EA-CG or KFI methods.

\subsubsection{Wall Clock Time}
For Cifar-10, Figure~\ref{fig:Nonconvex_cifar_time} shows that EA-CG method outperforms KFI method in terms of both training loss and testing accuracy. 
But the Fisher matrix with EA-CG method has the best performance. 
For ImageNet-10, Figure~\ref{fig:Nonconvex_imagenet_time} shows that EA-CG method surpasses KFI method regarding training loss.
But the performance of EA-CG and KFI methods is hard to distinguish by the testing accuracy.

\subsubsection{Epoch}
As shown in Figure~\ref{fig:Nonconvex_cifar_epoch}, for Cifar-10, the performance of EA-CG and KFI methods is hard to distinguish by the convergence speed of training loss in terms of epochs.
But the Fisher matrix with EA-CG method has the best training loss in Figure~\ref{fig:Nonconvex_cifar_epoch}.
Regarding testing accuracy, EA-CG method has the better performance than KFI method for this type of experiments.
For ImageNet-10, Figure~\ref{fig:Nonconvex_imagenet_epoch} shows that EA-CG method surpasses KFI method regarding training loss, which is consistent with Figure~\ref{fig:Nonconvex_imagenet_time}.
However, the performance of EA-CG and KFI methods is hard to distinguish by the testing accuracy, and the PCH matrix using KFI method has the best testing accuracy.

%table for Section 6-2
\begin{table}[t]
  \caption{Layer-wise errors between each of the approximate Hessian matrices and the true Hessian for the convex criterion function ``cross-entropy'' on ``Cifar-10''.}
  \label{table:Compare_to_True_Hessian}
  \centering
  \begin{tabular}{| c | c | c | c | c |}
    \hline
            & Fisher &   GN   &  PCH-1 & PCH-2 \\
    \hline
    Layer-1 & 0.0071 & 0.0057 & \bf{0.0036} & 0.0043 \\
    \hline
    Layer-2 & 0.0470 & \bf{0.0212} & 0.0237 & 0.0228 \\
    \hline
    Layer-3 & 0.1140 & 0.0220 & 0.0238 & \bf{0.0165} \\
    \hline
    Layer-4 & 0.0726 & 0.0207 & 0.0119 & \bf{0.0085} \\
    \hline
    Layer-5 & 0.0397 & 0.0130 & 0.0086 & \bf{0.0066} \\
    \hline
    Layer-6 & 0.0219 & 0.0107 & 0.0093 & \bf{0.0071} \\
    \hline
    Layer-7 & 0.0185 & 0.0176 & 0.0132 & \bf{0.0106} \\
    \hline
    Layer-8 & 0.0251 & \bf{0.0000} & \bf{0.0000} & \bf{0.0000} \\
    \hline
    Total   & 0.1535 & 0.0446 & 0.0402 & \bf{0.0330} \\
    \hline
  \end{tabular}
\end{table}

\subsection{Comparison with the True Hessian}
In addition to the comparison of training loss and testing accuracy on different criterion functions and datasets, we further examine the difference between the true Hessian and the approximate Hessian matrices such as the Fisher matrix.
Here, we measure the difference in the errors that is elaborated as follows.
Given the $t^{\text{th}}$ layer and the model parameters $\boldsymbol{\theta}$, the error is defined as 
\[
  \left\| \mathbb{E}_{i} [ \widetilde{\nabla_{\boldsymbol{b}^{t}}^{2}} \xi_{i}] -  |\mathbb{E}_{i} [\nabla_{\boldsymbol{b}^{t}}^{2} \xi_{i}]|   \right\|_{F},
\]
where $\widetilde{\nabla_{\boldsymbol{b}^{t}}^{2}} \xi_{i}$ stands for the approximate Hessian matrix, and $|\mathbb{E}_{i} [\nabla_{\boldsymbol{b}^{t}}^{2} \xi_{i}]|$ is to take the absolute values of the eigenvalues of $\mathbb{E}_{i} [\nabla_{\boldsymbol{b}^{t}}^{2} \xi_{i}]$, which we follow \cite{YND14a}.

The results are shown in Table~\ref{table:Compare_to_True_Hessian} where each value is derived from averaging the errors of the initial parameters $\boldsymbol{\theta}^{0}$ to the parameters $\boldsymbol{\theta}^{s-1}$ that are updated $s$ times.
Table~\ref{table:Compare_to_True_Hessian} reflects that the errors are not accumulated with layers for any approximate Hessian matrix.
This observation is important for KFRA and ours since both methods derive the curvature information by approximating the Hessian layer-by-layer recursively.
The ``Total'' row of Table~\ref{table:Compare_to_True_Hessian} also indicates that our proposed PCH-1 and PCH-2 are closer to the true Hessian than the Fisher and Gauss-Newton matrices.
The results of the other experiments echo our conclusions here and are provided in the Supplementary Materials A.

\section{Concluding Remarks}

To achieve more computationally feasible second-order methods for training FCNNs, we developed a practical approach, including our proposed PCH matrix and devised EA-CG method.
Our proposed PCH matrix overcomes the problem of training FCNNs with non-convex criterion functions.
Besides, EA-CG provides another alternative to efficiently derive update directions in this context.
Our empirical studies show that our proposed PCH matrix can compete with the state-of-the-art curvature approximation, and EA-CG does converge faster and enjoys better testing accuracy than KFI method.
Specially, the performance of our proposed approach is competitive with SGD.
As future work, we will extend the idea to work with convolutional nets.

\bibliographystyle{aaai}
\bibliography{swc}

%\section{Supplementary Materials }

\onecolumn

\clearpage
\section{Supplementary Materials A}
\subsection{Other True Hessian Comparison}
The other experiments of comparing the approximate Hessian matrices and the true Hessian matrix are reported in the Table~\ref{table:Compare_to_True_Hessian-extra-1}, Table~\ref{table:Compare_to_True_Hessian-extra-2}, and Table~\ref{table:Compare_to_True_Hessian-extra-3}.
All the three tables along with Table~\ref{table:Compare_to_True_Hessian} reflect that the errors are not accumulated with layers for any approximate Hessian matrix.
Because KFRA and our proposed method derive the curvature information by approximating the Hessian layer-by-layer recursively, we have to assert that the errors are not accumulated with layers.
Otherwise, the error accumulation will cause a severe problem in lower layers of FCNNs.
The consequence of these experiments provides an evidence to clear up this concern.
Moreover, the ``Total'' rows of these three tables also indicate that our proposed PCH-1 and PCH-2 are closer to the true Hessian than the Fisher and Gauss-Newton matrices.

\begin{table}[h]
  \centering
    \caption{Layer-wise errors between each of the approximate Hessian matrices and the true Hessian for the convex criterion function ``cross-entropy'' on ``ImageNet-10''.}
  %\caption{The average errors between each of approximate Hessian matrices and $\text{Pos-Eig}( \Expectation_{i} [\nabla_{\bb^{t}}^{2} \xi_{i}] )$, where the dataset, criterion function and activation function are ``ImageNet-10'', ``cross-entropy'' and ``sigmoid'', respectively.}
  \label{table:Compare_to_True_Hessian-extra-1}
  \begin{tabular}{| c | c | c | c | c |}
    \hline
            & Fisher &   GN   &  PCH-1 & PCH-2  \\
    \hline
    Layer-1 & 0.0012 & 0.0011 & \bf{0.0003} & 0.0009 \\
    \hline
    Layer-2 & 0.0240 & 0.0151 & 0.0146 & \bf{0.0132} \\
    \hline
    Layer-3 & 0.0409 & 0.0186 & 0.0198 & \bf{0.0174} \\
    \hline
    Layer-4 & 0.0390 & 0.0149 & 0.0137 & \bf{0.0160} \\
    \hline
    Layer-5 & 0.0327 & 0.0132 & 0.0115 & \bf{0.0112} \\
    \hline
    Layer-6 & 0.0279 & 0.0159 & 0.0142 & \bf{0.0128} \\
    \hline
    Layer-7 & 0.0290 & 0.0260 & 0.0259 & \bf{0.0183} \\
    \hline
    Layer-8 & 0.0427 & \bf{0.0000} & \bf{0.0000} & \bf{0.0000} \\
    \hline
    Total   & 0.0910 & 0.0436 & 0.0424 & \bf{0.0369} \\
    \hline
  \end{tabular}
  \caption{Layer-wise errors between each of the approximate Hessian matrices and the true Hessian for the non-convex criterion function ``Eq.~(\ref{eq:Non-Convex_Criterion})'' on ``Cifar-10''. The Gauss-Newton matrix is removed from this table since it is not PSD in this type of experiments.}
  %\caption{The average errors between each of approximate Hessian matrices and $\text{Pos-Eig}( \Expectation_{i} [\nabla_{\bb^{t}}^{2} \xi_{i}] )$, where the dataset, criterion function and activation function are ``Cifar-10'', ``Eq.~(\ref{eq:Non-Convex_Criterion})'' and ``sigmoid'', respectively.}
  \label{table:Compare_to_True_Hessian-extra-2}
  \begin{tabular}{| c | c | c | c | c |}
    \hline
            & Fisher & GN&  PCH-1 & PCH-2  \\
    \hline
    Layer-1 & 0.0044 & - & \bf{0.0035} & 0.0049 \\
    \hline
    Layer-2 & 0.0192 & - & \bf{0.0179} & 0.0456 \\
    \hline
    Layer-3 & 0.0528 & - & \bf{0.0185} & 0.0502 \\
    \hline
    Layer-4 & 0.0449 & - & \bf{0.0129} & 0.0344 \\
    \hline
    Layer-5 & 0.0330 & - & \bf{0.0092} & 0.0261 \\
    \hline
    Layer-6 & 0.0294 & - & \bf{0.0097} & 0.0198 \\
    \hline
    Layer-7 & 0.0331 & - & \bf{0.0140} & 0.0198 \\
    \hline
    Layer-8 & 0.0781 & - & \bf{0.0000} & \bf{0.0000} \\
    \hline
    Toatal  & 0.1198 & - & \bf{0.0349} & 0.0853 \\
    \hline
  \end{tabular}

  \caption{Layer-wise errors between each of the approximate Hessian matrices and the true Hessian for the non-convex criterion function ``Eq.~(\ref{eq:Non-Convex_Criterion})'' on ``ImageNet-10''. The Gauss-Newton matrix is removed from this table since it is not PSD in this type of experiments.}
  %\caption{The average errors between each of approximate Hessian matrices and $\text{Pos-Eig}( \Expectation_{i} [\nabla_{\bb^{t}}^{2} \xi_{i}] )$, where the dataset, criterion function and activation function are ``ImageNet-10'', ``Eq.~(\ref{eq:Non-Convex_Criterion})'' and ``sigmoid'', respectively.}
  \label{table:Compare_to_True_Hessian-extra-3}
  \begin{tabular}{| c | c | c | c | c |}
    \hline
            & Fisher & GN&  PCH-1 & PCH-2  \\
    \hline
    Layer-1 & 0.0003 & - & \bf{0.0002} & 0.0003 \\
    \hline
    Layer-2 & \bf{0.0116} & - & 0.0226 & 0.0253 \\
    \hline
    Layer-3 & \bf{0.0258} & - & 0.0264 & 0.0300 \\
    \hline
    Layer-4 & \bf{0.0248} & - & 0.0285 & 0.0326 \\
    \hline
    Layer-5 & \bf{0.0157} & - & 0.0284 & 0.0308 \\
    \hline
    Layer-6 & \bf{0.0147} & - & 0.0214 & 0.0227 \\
    \hline
    Layer-7 & 0.0219 & - & 0.0172 & \bf{0.0171} \\
    \hline
    Layer-8 & 0.0734 & - & \bf{0.0000} & \bf{0.0000} \\
    \hline
    Total   & 0.0880 & - & \bf{0.0598} & 0.0660 \\
    \hline
  \end{tabular}
\end{table}

\subsection{Comparison of different types of PCH matrices}
We conducted all the experiments considering two kinds of PCH matrices. However, their results are almost similar for all the experiments. For the sake of simplicity, we only show one of them in the figures of our experimental evaluation section and put their comparison here.

\begin{figure*}[h]
  \centering
    \includegraphics[width=0.4\columnwidth]{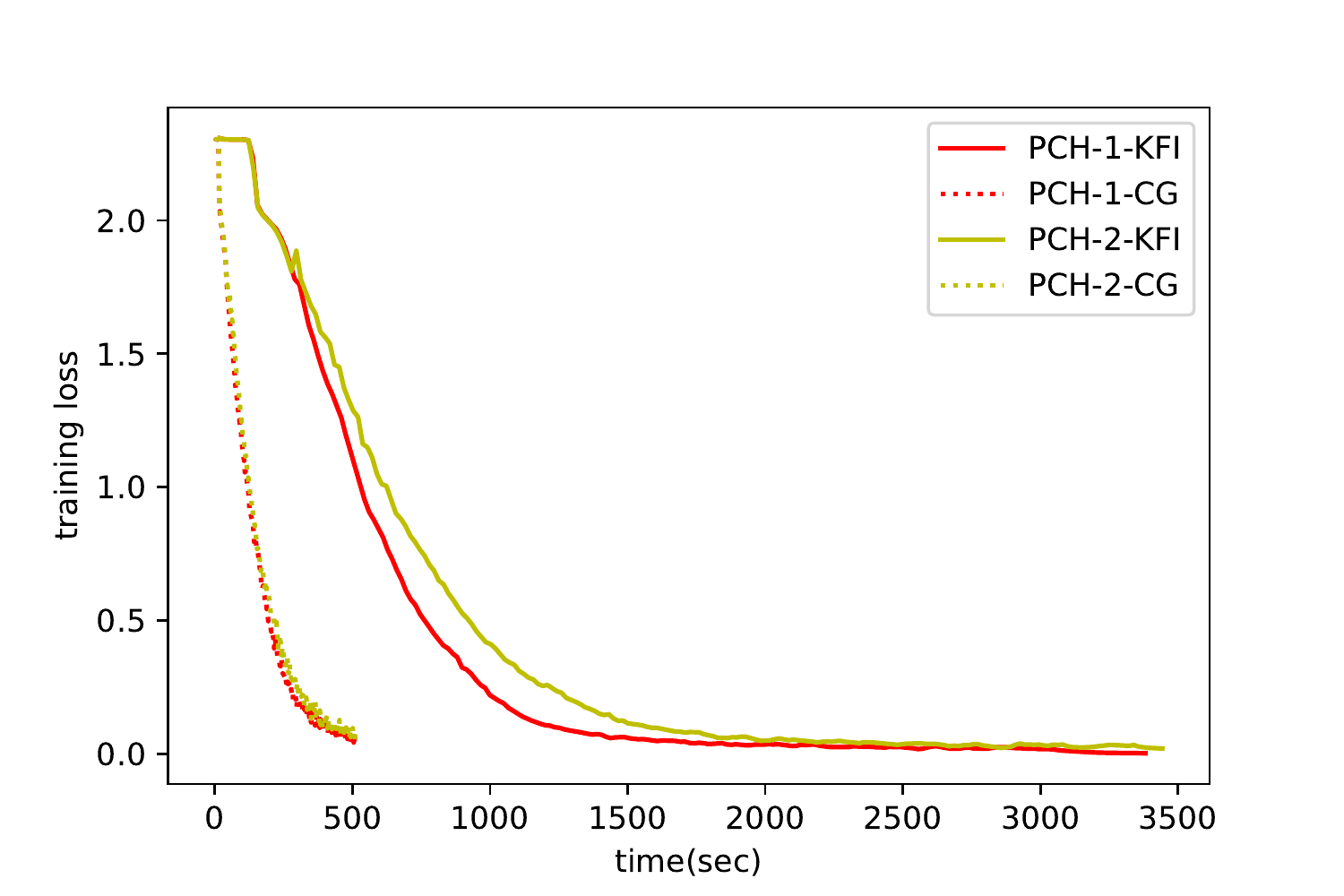}
    \includegraphics[width=0.4\columnwidth]{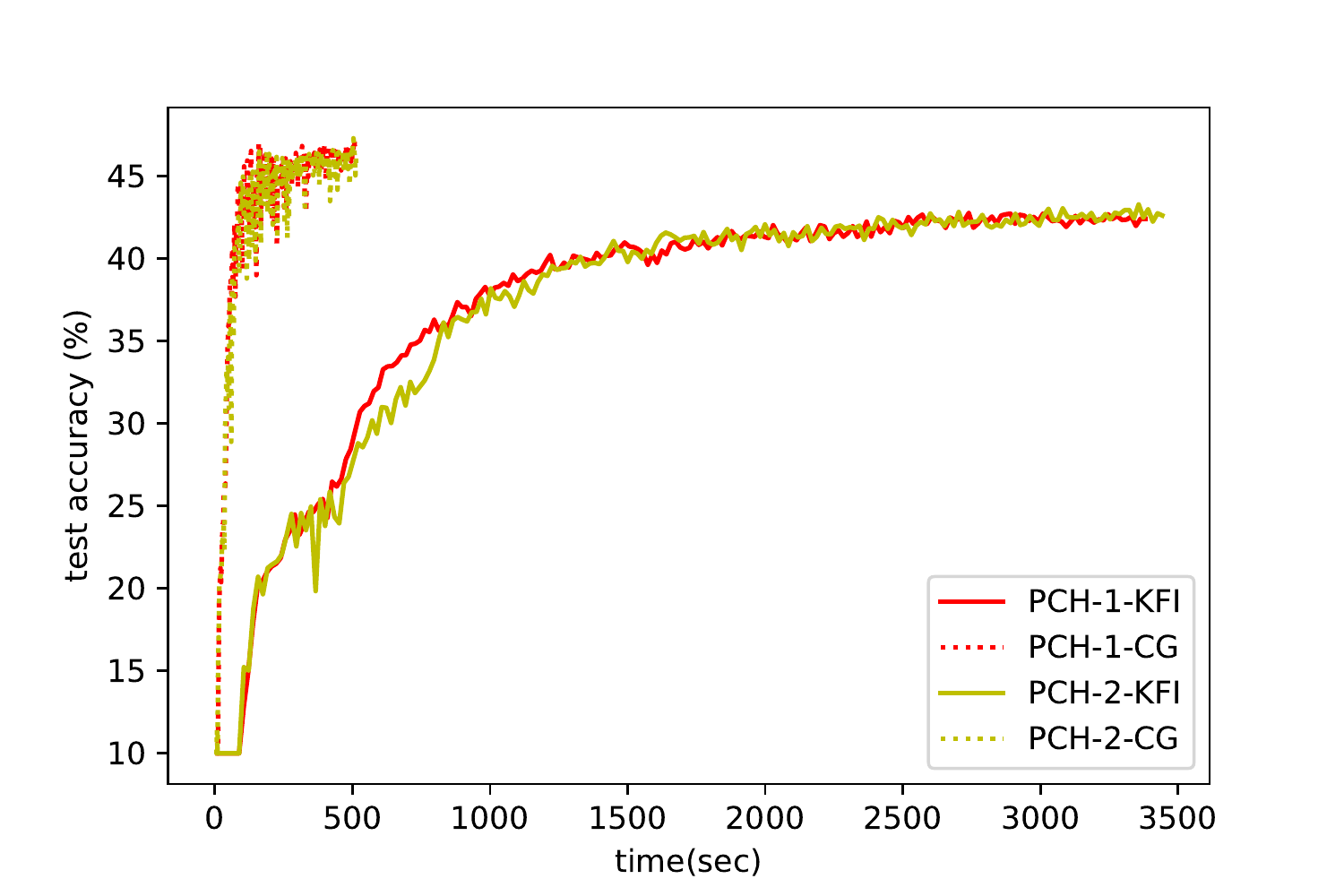}
  \caption{Comparison of PCH-1 and PCH-2 for convex criterion funciton ``cross-entropy'' on ``Cifar-10''.}
    \includegraphics[width=0.4\columnwidth]{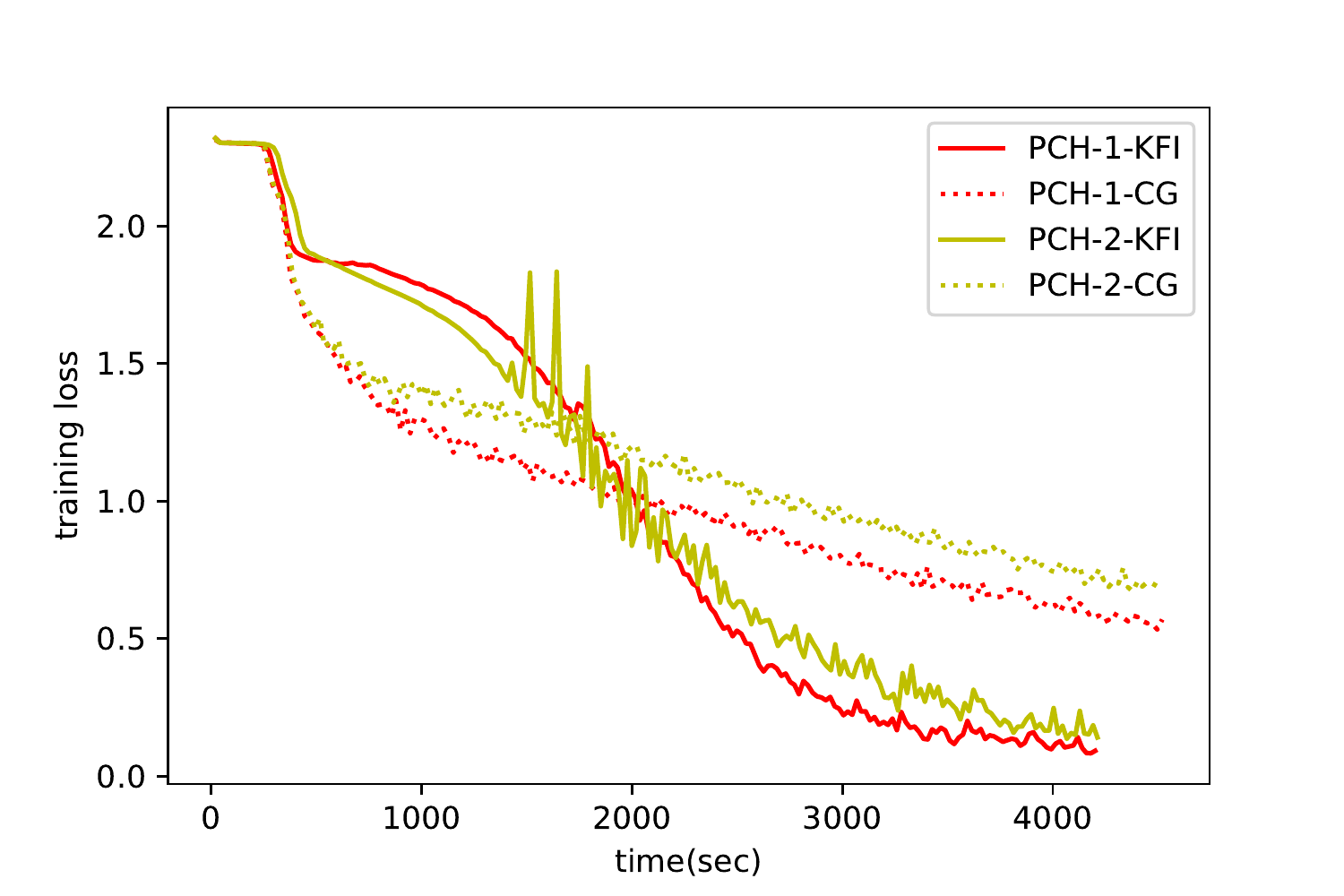}
    \includegraphics[width=0.4\columnwidth]{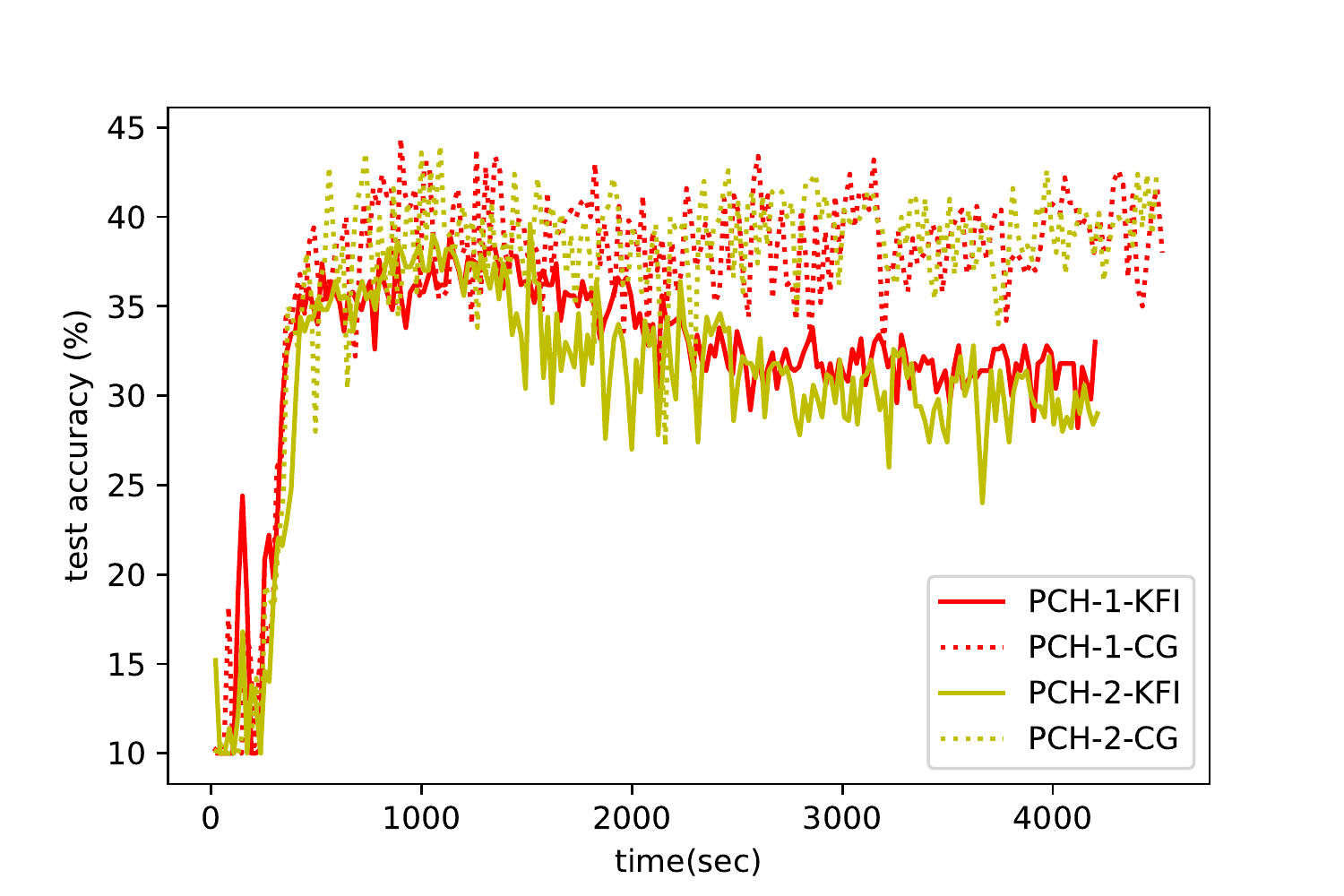}
  \caption{Comparison of PCH-1 and PCH-2 for convex criterion funciton ``cross-entropy'' on ``ImageNet-10''.}
    \includegraphics[width=0.4\columnwidth]{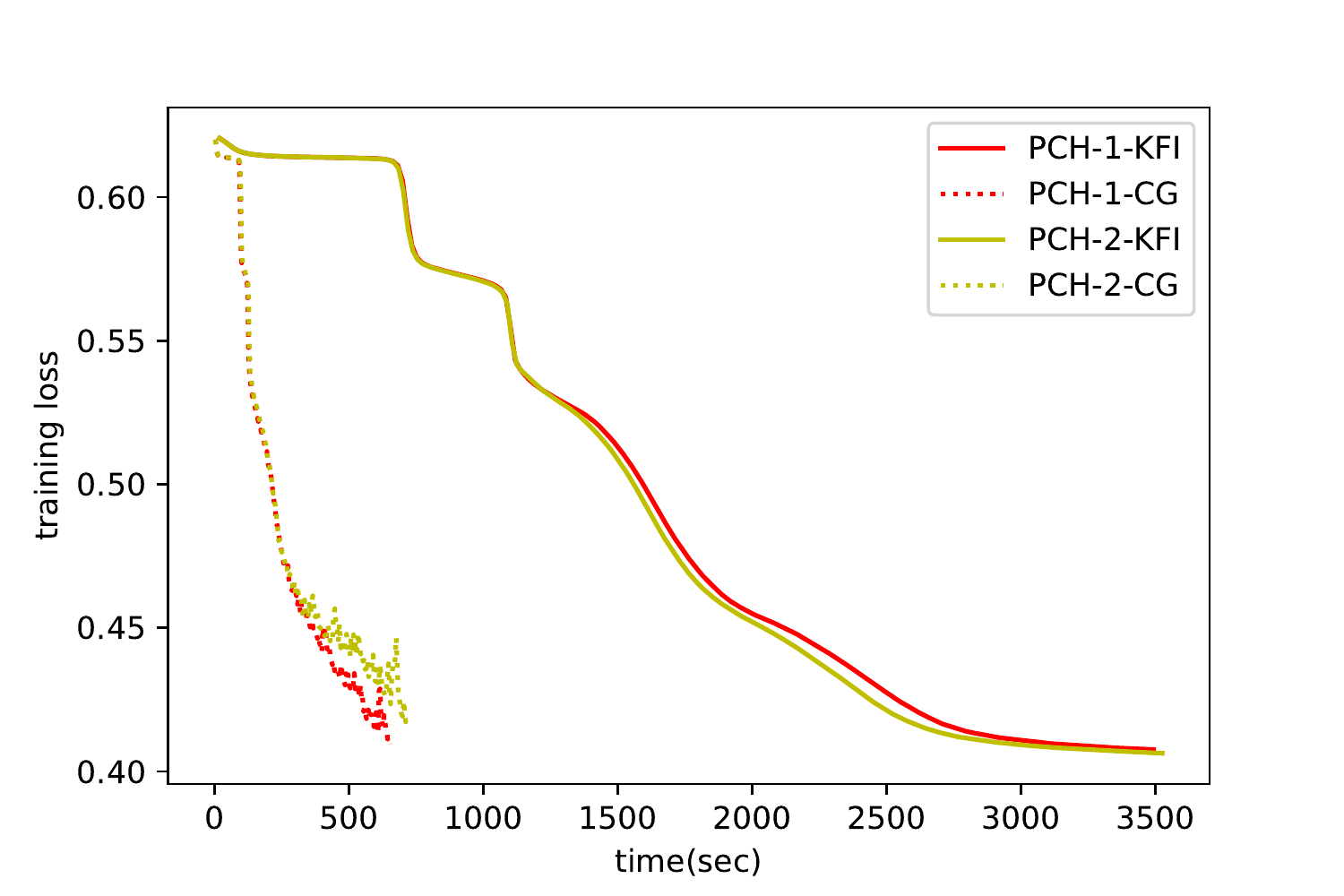}
    \includegraphics[width=0.4\columnwidth]{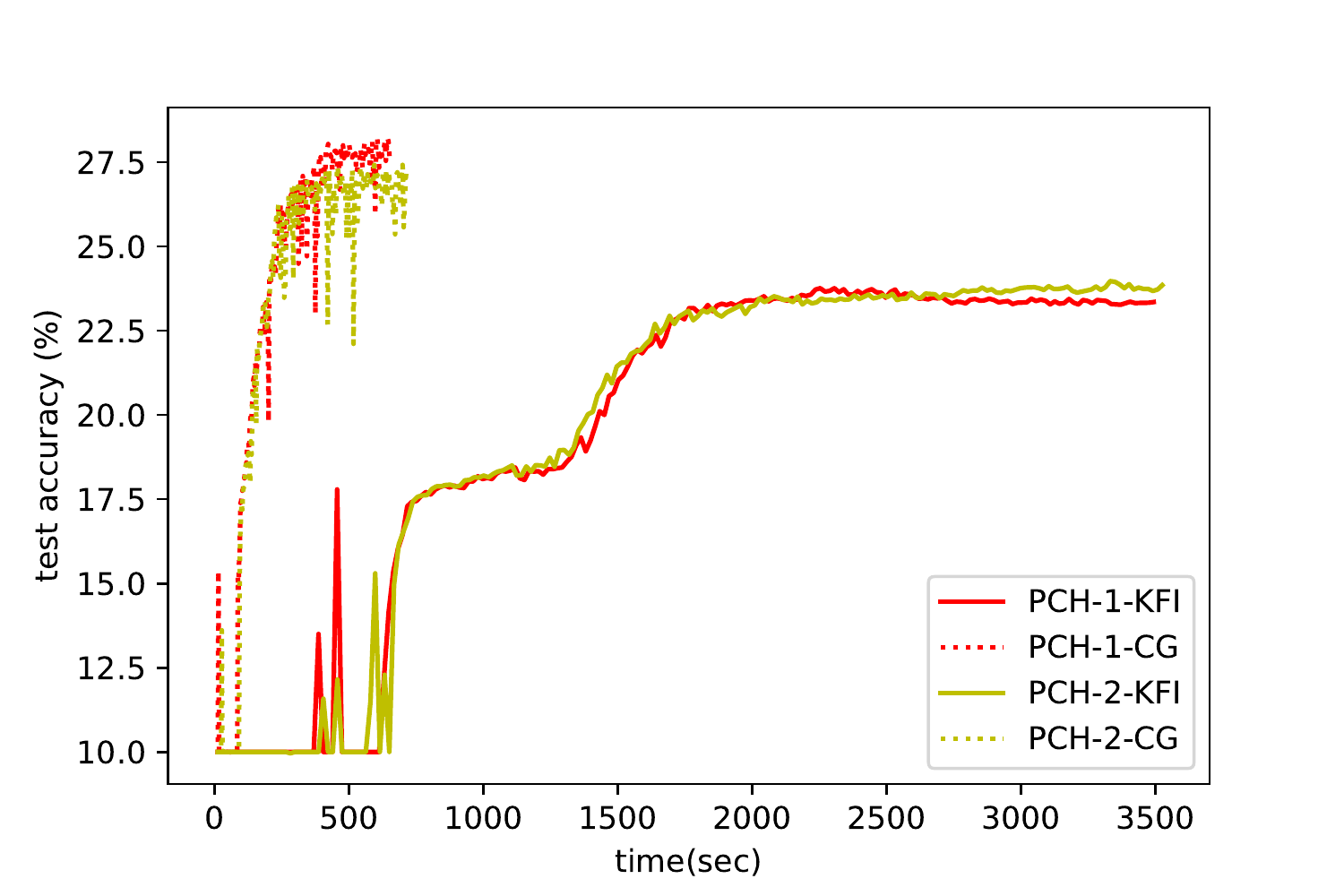}
  \caption{Comparison of PCH-1 and PCH-2 for non-convex criterion funciton ``Eq.~(\ref{eq:Non-Convex_Criterion})'' on ``Cifar-10''.}
    \includegraphics[width=0.4\columnwidth]{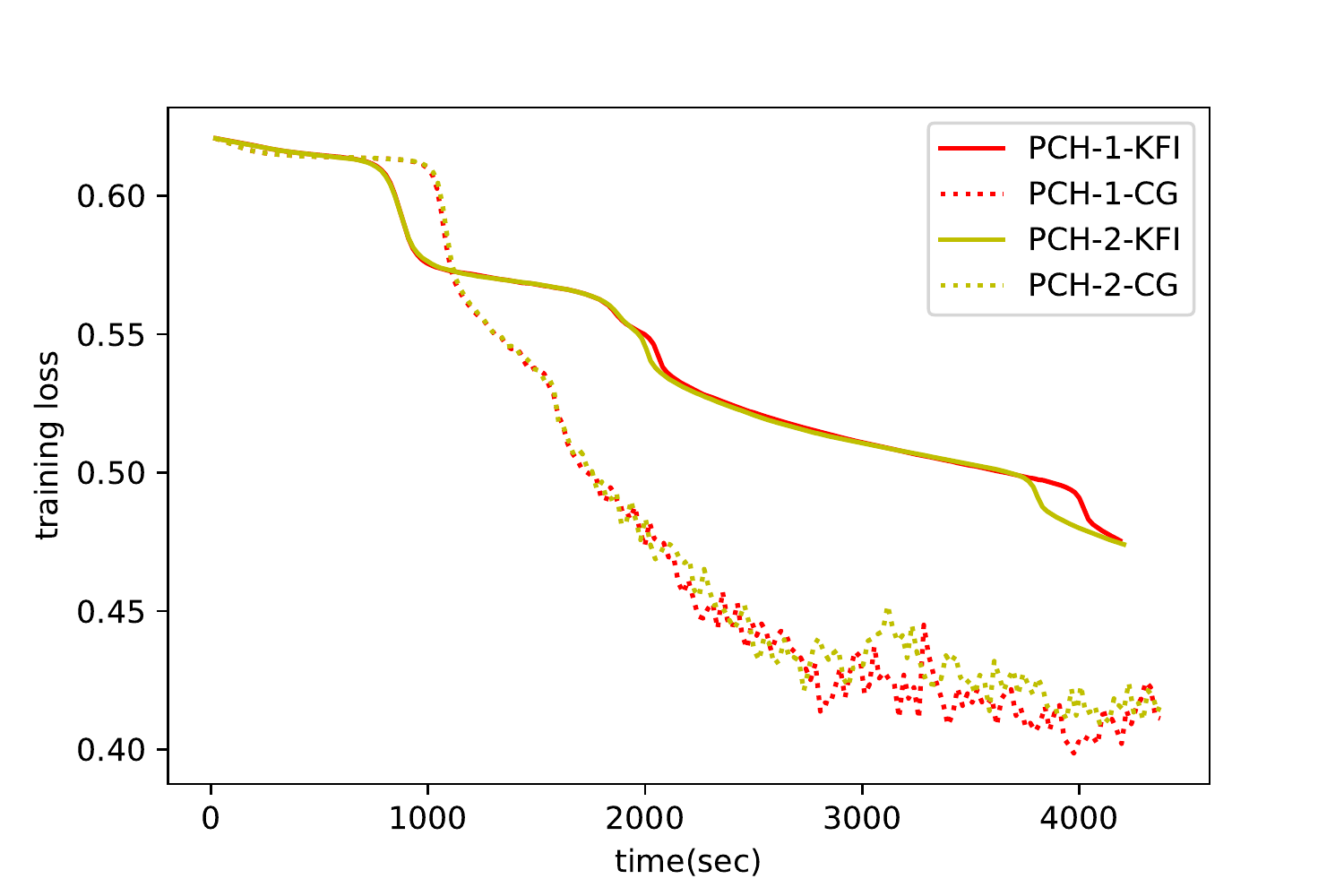}
    \includegraphics[width=0.4\columnwidth]{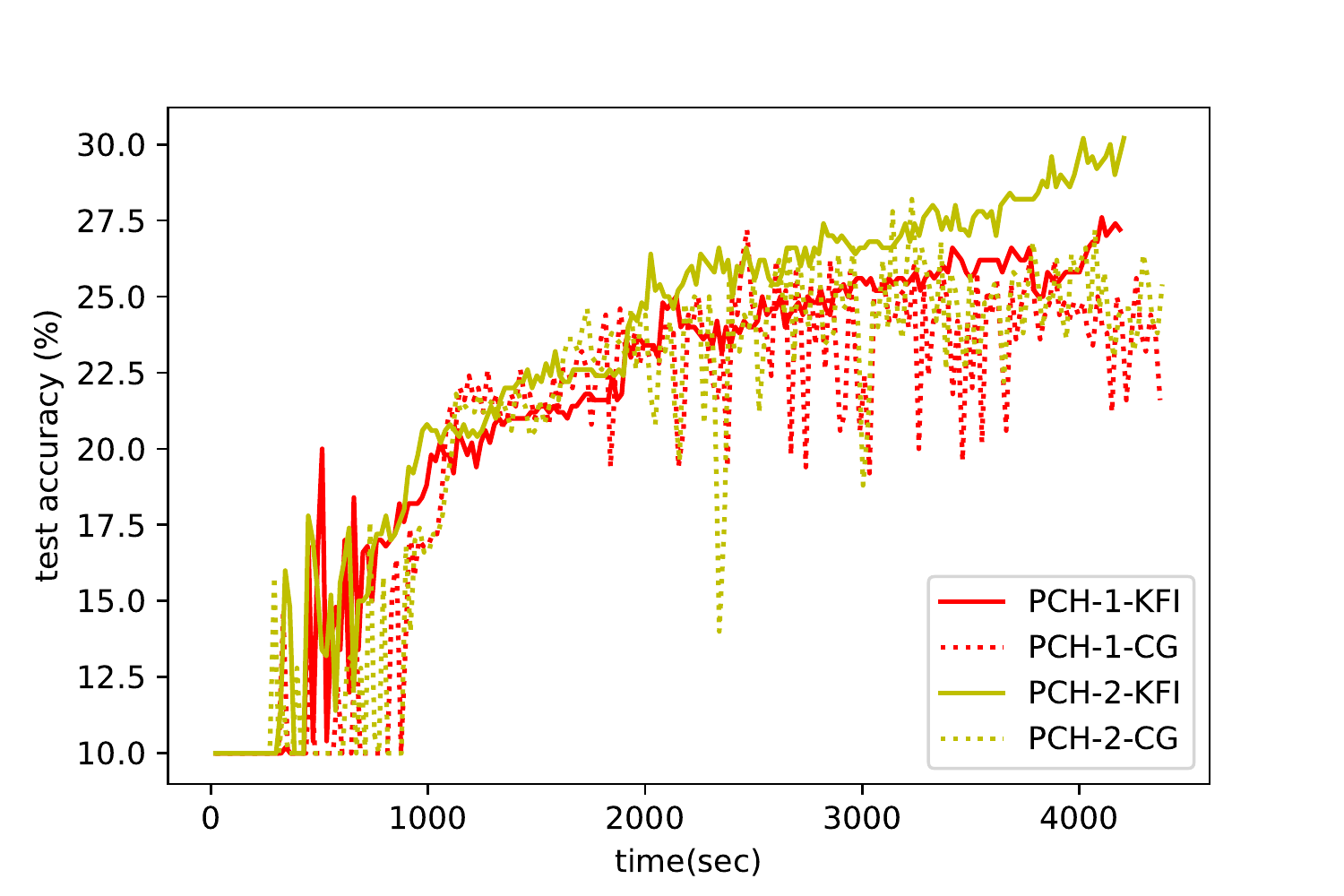}
  \caption{Comparison of PCH-1 and PCH-2 for non-convex criterion funciton ``Eq.~(\ref{eq:Non-Convex_Criterion})'' on ``ImageNet-10''.}
\end{figure*}

\clearpage
\section{Supplementary Materials B}
\begin{enumerate}[(i)]
\item $\nabla_{\bb^{k}}\xi_{i} = \nabla_{\hb_{i}^{k}}\xi_{i}$
  \begin{equation*}
  \begin{split}
       \frac{\partial \xi_{i} }{\partial \left[ \bb^{k}  \right]_{r} } 
    =& \sum_{s} \frac{\partial \xi_{i} }{\partial \left[ \hb_{i}^{k} \right]_{s}} 
       \frac{\partial \left[ \hb_{i}^{k} \right]_{s} }{\partial \left[ \bb^{k}  \right]_{r} } 
    = \sum_{s} \frac{\partial \xi_{i} }{\partial \left[ \hb_{i}^{k} \right]_{s}} 
		 \frac{\partial (\left[ \Wb^{k} \right]_{s \cdot } \hb_{i}^{k-1} + \left[ \bb^{k} \right]_{s}) }{\partial \left[ \bb^{k}  \right]_{r} }  \\
    =& \frac{\partial \xi_{i} }{\partial \left[ \hb_{i}^{k} \right]_{r}} 
       \frac{\partial (\left[ \Wb^{k} \right]_{r \cdot } \hb_{i}^{k-1} + \left[ \bb^{k} \right]_{r}) }{\partial \left[ \bb^{k}  \right]_{r} } 
    =  \frac{\partial \xi_{i} }{\partial \left[ \hb_{i}^{k}  \right]_{r} }. \\
  \end{split}
  \end{equation*}
  Therefore, we have $\nabla_{\bb^{k}}\xi_{i} = \nabla_{\hb_{i}^{k}}\xi_{i}$.

\item $\nabla_{\bb^{t-1}}\xi_{i} = \text{diag}( \hb_{i}^{(t-1)'}) \Wb^{tT} \nabla_{\bb^{t}}\xi_{i} , t = k, \ldots, 2$
\begin{enumerate}[{Case} 1:]
  \item $t = k$.
  \begin{equation*}
  \begin{split}
       \frac{\partial \xi_{i} }{\partial \left[ \bb^{k-1}  \right]_{r} } 
    =& \sum_{s} \frac{\partial \xi_{i} }{\partial \left[ \hb_{i}^{k} \right]_{s}} 
       \frac{\partial \left[ \hb_{i}^{k} \right]_{s} }{\partial \left[ \bb^{k-1}  \right]_{r} } 
    =  \sum_{s} \left\{ \frac{\partial \xi_{i} }{\partial \left[ \hb_{i}^{k} \right]_{s}} 
                        \sum_{u} \frac{\partial \left[ \hb_{i}^{k} \right]_{s} } {\partial \left[ \hb_{i}^{k-1} \right]_{u}}
                        \frac{\partial \left[ \hb_{i}^{k-1} \right]_{u}} {\partial \left[ \bb^{k-1}  \right]_{r} } \right\} \\
    =& \sum_{s} \left\{ \frac{\partial \xi_{i} }{\partial \left[ \hb_{i}^{k} \right]_{s}}
                        \sum_{u} \frac{\partial \left[ \hb_{i}^{k} \right]_{s} } {\partial \left[ \hb_{i}^{k-1} \right]_{u}}  
                        \frac{\partial \sigma (\left[ \Wb^{k-1} \right]_{u \cdot } \hb_{i}^{k-2} + \left[ \bb^{k-1} \right]_{u})} 
                             {\partial \left[ \bb^{k-1}  \right]_{r} } \right\} \\
    =& \sum_{s} \left\{ \frac{\partial \xi_{i} }{\partial \left[ \hb_{i}^{k} \right]_{s}} 
                        \frac{\partial \left[ \hb_{i}^{k} \right]_{s} } {\partial \left[ \hb_{i}^{k-1} \right]_{r}}
                        \frac{\partial \sigma (\left[ \Wb^{k-1} \right]_{r \cdot } \hb_{i}^{k-2} + \left[ \bb^{k-1} \right]_{r})} 
                             {\partial \left[ \bb^{k-1}  \right]_{r} } \right\} \\
    =& \sum_{s} \left\{ \frac{\partial \xi_{i} }{\partial \left[ \hb_{i}^{k} \right]_{s}} 
                        \frac{\partial \left[ \hb_{i}^{k} \right]_{s} } {\partial \left[ \hb_{i}^{k-1} \right]_{r}} 
                        \sigma ' (\left[ \Wb^{k-1} \right]_{r \cdot } \hb_{i}^{k-2} + \left[ \bb^{k-1} \right]_{r}) \right\}  \\
    =& \sum_{s} \left\{ \frac{\partial \xi_{i} }{\partial \left[ \hb_{i}^{k} \right]_{s}} 
                        \frac{\partial \left[ \hb_{i}^{k} \right]_{s} } {\partial \left[ \hb_{i}^{k-1} \right]_{r}}
                        \left[ \hb_{i}^{(k-1)'} \right]_{r} \right\} \\
    =& \sum_{s} \left\{ \frac{\partial \xi_{i} }{\partial \left[ \hb_{i}^{k} \right]_{s}} 
                        \frac{\partial (\left[ \Wb^{k} \right]_{s \cdot } \hb_{i}^{k-1} + \left[ \bb^{k} \right]_{s}) } 
                             {\partial \left[ \hb_{i}^{k-1} \right]_{r}} 
                        \left[ \hb_{i}^{(k-1)'} \right]_{r} \right\} \\
    =& \sum_{s} \left\{ \frac{\partial \xi_{i} }{\partial \left[ \hb_{i}^{k} \right]_{s}} 
                        \left[ \Wb^{k} \right]_{s r} \left[ \hb_{i}^{(k-1)'} \right]_{r} \right\} 
    =  \left[ \hb_{i}^{(k-1)'} \right]_{r}  {\nabla_{\hb_{i}^{k}} \xi_{i}}^{T} \left[ \Wb^{k} \right]_{\cdot r} \\
    =& \left[ \hb_{i}^{(k-1)'} \right]_{r} {\nabla_{\bb^{k}} \xi_{i}}^{T} \left[ \Wb^{k} \right]_{\cdot r} 
    = \left[ \hb_{i}^{(k-1)'} \right]_{r} \left[ \Wb^{kT} \right]_{r \cdot}\nabla_{\bb^{k}} \xi_{i}. \\
  \end{split}
  \end{equation*}
  Therefore, we have $\nabla_{\bb^{k-1}}\xi_{i} = \text{diag}( \hb_{i}^{(k-1)T} ) \Wb^{kT} \nabla_{\bb^{k}}\xi_{i}$.
      
  \item $t < k$. Similarly, 
  \begin{equation*}
  \begin{split}
       \frac{\partial \xi_{i} }{\partial \left[ \bb^{t-1}  \right]_{r} } 
    =& \sum_{s} \left\{ \frac{\partial \xi_{i}}{\partial \left[ \hb_{i}^{t} \right]_{s}} 
                        \frac{\partial \left[ \hb_{i}^{t} \right]_{s} } {\partial \left[ \hb_{i}^{t-1} \right]_{r}}
                        \left[ \hb_{i}^{(t-1)'} \right]_{r} \right\} \\
    =& \sum_{s} \left\{ \frac{\partial \xi_{i}}{\partial \left[ \hb_{i}^{t} \right]_{s}} 
                        \frac{\partial \sigma (\left[ \Wb^{t} \right]_{s \cdot } \hb_{i}^{t-1} + \left[ \bb^{t} \right]_{s}) } 
                             {\partial \left[ \hb_{i}^{t-1} \right]_{r}} 
                        \left[ \hb_{i}^{(t-1)'} \right]_{r} \right\} \\
    =& \sum_{s} \left\{ \frac{\partial \xi_{i}}{\partial \left[ \hb_{i}^{t} \right]_{s}} 
                        \sigma'(\left[ \Wb^{t} \right]_{s \cdot } \hb_{i}^{t-1} + \left[ \bb^{t} \right]_{s}) 
                        \left[ \Wb^{t} \right]_{s r } \left[ \hb_{i}^{(t-1)'} \right]_{r} \right\} \\
    =& \sum_{s} \left\{ \frac{\partial \xi_{i}}{\partial \left[ \hb_{i}^{t} \right]_{s}} 
                        \frac{\partial \sigma (\left[ \Wb^{t} \right]_{s \cdot } \hb_{i}^{t-1} + \left[ \bb^{t} \right]_{s}) } 
                             {\partial \left[ \bb^{t} \right]_{s}} \left[ \Wb^{t} \right]_{s r }      
                        \left[ \hb_{i}^{(t-1)'} \right]_{r} \right\}  \\
    =& \sum_{s} \left\{ \frac{\partial \xi_{i}}{\partial \left[ \hb_{i}^{t} \right]_{s}} 
                        \frac{\partial \left[ \hb_{i}^{t} \right]_{s} } {\partial \left[ \bb^{t} \right]_{s}}
                        \left[ \Wb^{t} \right]_{s r }      
                        \left[ \hb_{i}^{(t-1)'} \right]_{r} \right\} \\
    =& \sum_{s} \left\{ \sum_{u} \left\{ \frac{\partial \xi_{i}}{\partial \left[ \hb_{i}^{t} \right]_{u}} 
                                         \frac{\partial \left[ \hb_{i}^{t} \right]_{u} } {\partial \left[ \bb^{t} \right]_{s}}\right\}
                                         \left[ \Wb^{t} \right]_{s r } \left[ \hb_{i}^{(t-1)'} \right]_{r} \right\} \\
    =& \sum_{s} \left\{ \frac{\partial \xi_{i}} {\partial \left[ \bb^{t} \right]_{s}}
                        \left[ \Wb^{t} \right]_{s r } \left[ \hb_{i}^{(t-1)'} \right]_{r} \right\}
    = \left[ \hb_{i}^{(t-1)'} \right]_{r}  \left[ \Wb^{tT} \right]_{r \cdot}\nabla_{\bb^{t}} \xi_{i}. \\
  \end{split}
  \end{equation*}
  Therefore, we have $ \nabla_{\bb^{t-1}}\xi_{i} = \text{diag}( \hb_{i}^{(t-1)T} )\Wb^{tT} \nabla_{\bb^{t}}\xi_{i}$.
\end{enumerate}

\item $\nabla_{\Wb^{t}}\xi_{i} = \nabla_{\bb^{t}}\xi_{i} \otimes \hb_{i}^{(t-1)T}, t = k , \ldots, 1 $
\begin{enumerate}[{Case} 1:]
  \item $t = k$.
  \begin{equation*}
  \begin{split}
       \frac{\partial \xi_{i} }{\partial \left[ \Wb^{k} \right]_{\alpha \beta}} 
    =& \sum_{s} \frac{\partial \xi_{i} } { \partial \left[ \hb_{i}^{k} \right]_{s} }
                \frac{\partial \left[ \hb_{i}^{k} \right]_{s} } {\partial \left[ \Wb^{k} \right]_{\alpha \beta}} 
    =  \sum_{s} \frac{\partial \xi_{i} } { \partial \left[ \hb_{i}^{k} \right]_{s} }
                \frac{\partial (\left[ \Wb^{k} \right]_{s \cdot} \hb_{i}^{k-1} + \left[ \bb^{k} \right]_{s})} 
                     {\partial \left[ \Wb^{k} \right]_{\alpha \beta}} \\
    =& \frac{\partial \xi_{i} } { \partial \left[ \hb_{i}^{k} \right]_{\alpha} }
       \frac{\partial (\left[ \Wb^{k} \right]_{\alpha \cdot} \hb_{i}^{k-1} + \left[ \bb^{k} \right]_{\alpha})} 
            {\partial \left[ \Wb^{k} \right]_{\alpha \beta}} \\
    =& \frac{\partial \xi_{i} } { \partial \left[ \hb_{i}^{k} \right]_{\alpha} }
       \frac{\partial (\left[ \Wb^{k} \right]_{\alpha \cdot} \hb_{i}^{k-1} + \left[ \bb^{k} \right]_{\alpha})} 
            {\partial \left[ \Wb^{k} \right]_{\alpha \beta}} \\
    =& \frac{\partial \xi_{i} } { \partial \left[ \hb_{i}^{k} \right]_{\alpha} } \left[ \hb_{i}^{k-1} \right]_{\beta} 
    =  \frac{\partial \xi_{i} } { \partial \left[ \bb_{i}^{k} \right]_{\alpha} } \left[ \hb_{i}^{k-1} \right]_{\beta}. \\
  \end{split}
  \end{equation*}
  Therefore, we have $\nabla_{\Wb^{k}}\xi_{i} = \nabla_{\bb^{k}}\xi_{i} \otimes \hb_{i}^{(k-1)T}$. 
      
  \item $t < k$. 
  Similarly,
  \begin{equation*}
  \begin{split}
       \frac{\partial \xi_{i} }{\partial \left[ \Wb^{t} \right]_{\alpha \beta}} 
    =& \sum_{s} \frac{\partial \xi_{i} } { \partial \left[ \hb_{i}^{t} \right]_{s} }
                \frac{\partial \sigma (\left[ \Wb^{t} \right]_{s \cdot} \hb_{i}^{t-1} + \left[ \bb^{t} \right]_{s})} 
                     {\partial \left[ \Wb^{t} \right]_{\alpha \beta}}  \\
    =& \frac{\partial \xi_{i} } { \partial \left[ \hb_{i}^{t} \right]_{\alpha} }
       \sigma'(\left[ \Wb^{t} \right]_{\alpha \cdot} \hb_{i}^{t-1} + \left[ \bb^{t} \right]_{\alpha})
       \left[ \hb_{i}^{t-1} \right]_{\beta} 
    = \frac{\partial \xi_{i} } { \partial \left[ \hb_{i}^{t} \right]_{\alpha} }
       \frac{\partial \sigma (\left[ \Wb^{t} \right]_{\alpha \cdot} \hb_{i}^{t-1} + \left[ \bb^{t} \right]_{\alpha})}
            {\partial \left[ \bb^{t} \right]_{\alpha}}                 
       \left[ \hb_{i}^{t-1} \right]_{\beta} \\
    =& \frac{\partial \xi_{i} } { \partial \left[ \hb_{i}^{t} \right]_{\alpha} }
       \frac{\partial \left[ \hb_{i}^{t} \right]_{\alpha}} {\partial \left[ \bb^{t} \right]_{\alpha}}                 
       \left[ \hb_{i}^{t-1} \right]_{\beta}  
    = \sum_{s} \left\{ \frac{\partial \xi_{i} } { \partial \left[ \hb_{i}^{t} \right]_{s} }
                       \frac{\partial \left[ \hb_{i}^{t} \right]_{s}} {\partial \left[ \bb^{t} \right]_{\alpha}} \right\}                 
               \left[ \hb_{i}^{t-1} \right]_{\beta} \\
    =& \frac{\partial \xi_{i} } {\partial \left[ \bb^{t} \right]_{\alpha}}                 
       \left[ \hb_{i}^{t-1} \right]_{\beta} 
    =  \left[ \nabla_{\bb^{t}}\xi_{i} \right]_{\alpha}
       \left[ \hb_{i}^{t-1} \right]_{\beta}.  \\
	\end{split}
	\end{equation*}
  Therefore, we have $\nabla_{\Wb^{t}}\xi_{i} = \nabla_{\bb^{t}}\xi_{i} \otimes \hb_{i}^{(t-1)T}$. 
\end{enumerate}

\end{enumerate}

\clearpage
\section{Supplementary Materials C}
\begin{enumerate}[(i)]
\item $\nabla_{\bb^{k}}^{2} \xi_{i} = \nabla_{\hb_{i}^{k}}^{2} \xi_{i} $
  \begin{equation*}
  \begin{split}
       \frac{\partial^{2} \xi_{i} }{\partial \left[ \bb^{k} \right]_{r} \partial \bb^{k} } 
    =& \frac{\partial \nabla_{\bb^{k}} \xi_{i}}{\partial \left[ \bb^{k} \right]_{r} } 
    =  \frac{\partial \nabla_{\hb_{i}^{k}} \xi_{i}}{\partial \left[ \bb^{k} \right]_{r} } 
    =  \sum_{s} \frac{\partial \nabla_{\hb_{i}^{k}} \xi_{i}} {\partial \left[ \hb_{i}^{k} \right]_{s} }
                \frac{\partial \left[ \hb_{i}^{k} \right]_{s} } {\partial \left[ \bb^{k} \right]_{r} } \\
    =& \sum_{s} \frac{\partial \nabla_{\hb_{i}^{k}} \xi_{i}} {\partial \left[ \hb_{i}^{k} \right]_{s} }
                \frac{\partial (\left[ \Wb^{k} \right]_{s \cdot} \hb_{i}^{k-1} + \left[ \bb^{k} \right]_{s}) } {\partial \left[ \bb^{k} \right]_{r} } 
    = \frac{\partial \nabla_{\hb_{i}^{k}} \xi_{i}} {\partial \left[ \hb_{i}^{k} \right]_{r} }. \\
  \end{split}
  \end{equation*}
  Therefore, we have $\nabla_{\bb^{k}}^{2} \xi_{i} = \nabla_{\hb_{i}^{k}}^{2} \xi_{i}$.

\item $ \nabla_{\bb^{t-1}}^{2} \xi_{i} = \text{diag}(\hb_{i}^{(t-1)'})\Wb^{tT} \nabla_{\bb^{t}}^{2} \xi_{i} \Wb^{t} \text{diag}(\hb_{i}^{(t-1)'})
                                         + \text{diag}(\hb_{i}^{(t-1)''} \odot (\Wb^{tT} \nabla_{\bb^{t}} \xi_{i})),$ \par
      $t = k, \ldots, 2$

\begin{enumerate}[{Case} 1:]
  \item $t = k$.
  \begin{equation*}
  \begin{split}
       \frac{\partial^{2} \xi_{i} }{\partial \left[ \bb^{k-1} \right]_{r} \partial \bb^{k-1} } 
    =& \frac{\partial \nabla_{\bb^{k-1}}\xi_{i} } {\partial \left[ \bb^{k-1} \right]_{r} } 
    =  \frac{\partial \text{ diag}(\hb_{i}^{(k-1)'}) \Wb^{kT} \nabla_{\bb^{k}}\xi_{i} } {\partial \left[ \bb^{k-1} \right]_{r} } \\
    =& \underset{\text{Part 1}}{ \underbrace{
       \frac{\partial \text{ diag}(\hb_{i}^{(k-1)'}) \Wb^{kT} } {\partial \left[ \bb^{k-1} \right]_{r} }  \nabla_{\bb^{k}}\xi_{i} }} 
     +\underset{\text{Part 2}}{ \underbrace{ 
      \text{diag}(\hb_{i}^{(k-1)'}) \Wb^{kT} \frac{ \partial\nabla_{\bb^{k}}\xi_{i} } {\partial \left[ \bb^{k-1} \right]_{r} }  }}. \\
  \end{split}
  \end{equation*}
  %Part 1
  \begin{equation*}
  \begin{split}
       \text{Part 1}
    =& r^{\text{th}} \leftarrow 
       \begin{bmatrix}
         0 \\ 
         \vdots \\
         0 \\
         \left[ \hb_{i}^{(k-1)''} \right]_{r} \left[ \Wb^{kT} \right]_{r \cdot} \nabla_{\bb^{k}}\xi_{i} \\
         0 \\
         \vdots \\
         0 \end{bmatrix} 
    = ( \left[ \hb_{i}^{(k-1)''} \right]_{r} \left[ \Wb^{kT} \right]_{r \cdot} \nabla_{\bb^{k}}\xi_{i} ) \eb_{r}. \\
  \end{split}
  \end{equation*}
  Next, we derive
  %Part 2
  \begin{equation*}
  \begin{split}
       \frac{\partial \nabla_{\bb^{k}} \xi_{i} } {\partial \left[ \bb^{k-1} \right]_{r} } 
    =& \sum_{s} \frac{\partial \nabla_{\bb^{k}} \xi_{i} } {\partial \left[ \hb_{i}^{k} \right]_{s} }  
                \frac{\partial \left[ \hb_{i}^{k} \right]_{s} }{\partial \left[ \bb^{k-1} \right]_{r} } 
    =  \sum_{s} \frac{\partial \nabla_{\bb^{k}} \xi_{i} } {\partial \left[ \hb_{i}^{k} \right]_{s} }  
       \sum_{u} \frac{\partial \left[ \hb_{i}^{k} \right]_{s} } {\partial \left[ \hb_{i}^{k-1} \right]_{u} }
                \frac{\partial \left[ \hb_{i}^{k-1} \right]_{u} } {\partial \left[ \bb^{k-1} \right]_{r} } \\
    =& \sum_{s} \frac{\partial \nabla_{\bb^{k}} \xi_{i} } {\partial \left[ \hb_{i}^{k} \right]_{s} }  
                \frac{\partial \left[ \hb_{i}^{k} \right]_{s} } {\partial \left[ \hb_{i}^{k-1} \right]_{r} }
                \left[ \hb_{i}^{(k-1)'} \right]_{r} \\
    =& \sum_{s} \frac{\partial \nabla_{\bb^{k}} \xi_{i} } {\partial \left[ \hb_{i}^{k} \right]_{s} }  
                \frac{\partial (\left[ \Wb^{k} \right]_{s \cdot} \hb_{i}^{k-1} + \left[ \bb^{k} \right]_{s}) } {\partial \left[ \hb_{i}^{k-1} \right]_{r} } 
                \left[ \hb_{i}^{(k-1)'} \right]_{r} \\
   = &\sum_{s} \frac{\partial \nabla_{\bb^{k}} \xi_{i} } {\partial \left[ \hb_{i}^{k} \right]_{s} }  
               \left[ \Wb^{k} \right]_{s r} \left[ \hb_{i}^{(k-1)'} \right]_{r} 
   =  \sum_{s} \frac{\partial \nabla_{\bb^{k}} \xi_{i} } {\partial \left[ \bb_{i}^{k} \right]_{s} }  
               \left[ \Wb^{k} \right]_{s r} \left[ \hb_{i}^{(k-1)'} \right]_{r} \\
   = &\sum_{s} \left[ \nabla_{\bb^{k}}^{2} \xi_{i} \right]_{\cdot s} \left[ \Wb^{k} \right]_{s r} \left[ \hb_{i}^{(k-1)'} \right]_{r} 
   =  \nabla_{\bb^{k}}^{2} \xi_{i} \left[ \Wb^{k} \right]_{\cdot r} \left[ \hb_{i}^{(k-1)'} \right]_{r}. \\
  \end{split}
  \end{equation*}
  Therefore, we have
  $ \text{Part 2} = \text{diag}(\hb_{i}^{(k-1)'})\Wb^{kT} \nabla_{\bb^{k}}^{2} \xi_{i} \left[ \Wb^{k} \text{diag}( \hb_{i}^{(k-1)'} ) \right]_{\cdot r}$, then
  \begin{equation*}
  \begin{split}
      \nabla_{\bb^{k-1}}^{2} \xi_{i} 
    =& \text{diag}(\hb_{i}^{(k-1)'} ) \Wb^{kT} \nabla_{\bb^{k}}^{2} \xi_{i}  \Wb^{k}  \text{diag}(\hb_{i}^{(k-1)'} ) \\
     & + \text{diag}(\hb_{i}^{(k-1)''}  \odot (\Wb^{kT} \nabla_{\bb^{k}} \xi_{i})).\\
  \end{split}
  \end{equation*}

  \item $t < k$. 
  Similarly, 
  \begin{equation*}
      \frac{\partial^{2} \xi_{i} }{\partial \left[ \bb^{t-1} \right]_{r} \partial \bb^{t-1} } 
    = \underset{\text{Part 1}}
               {\underbrace{ \frac{\partial \text{ diag}(\hb_{i}^{(t-1)'} )\Wb^{tT}}
                                  {\partial \left[ \bb^{t-1} \right]_{r} } \nabla_{\bb^{t}}\xi_{i} } } 
     +\underset{\text{Part 2}}
               {\underbrace{ \text{diag}(\hb_{i}^{(t-1)'} )\Wb^{tT} \frac{ \partial\nabla_{\bb^{t}}\xi_{i} } 
                                                                         {\partial \left[ \bb^{t-1} \right]_{r} }  }}.
  \end{equation*}
  \[
        \text{Part 1} = ( \left[ \hb_{i}^{(t-1)''} \right]_{r} \left[ \Wb^{tT} \right]_{r \cdot} \nabla_{\bb^{t}}\xi_{i} ) \eb_{r},
  \]
  and
  \begin{equation*}
  \begin{split}
        \frac{\partial \nabla_{\bb^{t}} \xi_{i} } {\partial \left[ \bb^{t-1} \right]_{r} } 
    =& \sum_{s} \frac{\partial \nabla_{\bb^{t}} \xi_{i} } {\partial \left[ \hb_{i}^{t} \right]_{s} }  
                \frac{\partial \sigma(\left[ \Wb^{t} \right]_{s \cdot} \hb_{i}^{t-1} + \left[ \bb^{t} \right]_{s})}
                     {\partial \left[ \hb_{i}^{t-1} \right]_{r} } 
                \left[ \hb_{i}^{(t-1)'} \right]_{r} \\
    =& \sum_{s} \frac{\partial \nabla_{\bb^{t}} \xi_{i} } {\partial \left[ \hb_{i}^{t} \right]_{s} }
                \left[ \hb_{i}^{(t)'} \right]_{s} \left[ \Wb^{t} \right]_{s r} \left[ \hb_{i}^{(t-1)'} \right]_{r} \\
    =& \sum_{s} \frac{\partial \nabla_{\bb^{t}} \xi_{i} } {\partial \left[ \hb_{i}^{t} \right]_{s} }
                \frac{\partial \left[ \hb_{i}^{t} \right]_{s}}{\partial \left[ \bb^{t} \right]_{s}}
                \left[ \Wb^{t} \right]_{s r} \left[ \hb_{i}^{(t-1)'} \right]_{r} \\
    =& \sum_{s} \sum_{u} \left\{ \frac{\partial \nabla_{\bb^{t}} \xi_{i} } {\partial \left[ \hb_{i}^{t} \right]_{u} }
                         \frac{\partial \left[ \hb_{i}^{t} \right]_{u}}{\partial \left[ \bb^{t} \right]_{s}} \right\}
                         \left[ \Wb^{t} \right]_{s r} \left[ \hb_{i}^{(t-1)'} \right]_{r} \\
    =& \sum_{s} \frac{\partial \nabla_{\bb^{t}} \xi_{i} } {\partial \left[ \bb_{i}^{t} \right]_{s} }  
                \left[ \Wb^{t} \right]_{s r} \left[ \hb_{i}^{(t-1)'} \right]_{r} 
    =  \sum_{s} \left[ \nabla_{\bb^{t}}^{2} \xi_{i} \right]_{\cdot s} \left[ \Wb^{t} \right]_{s r} \left[ \hb_{i}^{(t-1)'} \right]_{r} \\
    = &  \nabla_{\bb^{t}}^{2} \xi_{i} \left[ \wb^{t} \right]_{\cdot r} \left[ \hb_{i}^{(t-1)'} \right]_{r}. \\
  \end{split}
  \end{equation*}      
  \end{enumerate}
  By combining the case 1 and 2, we have
  \begin{equation*}
      \nabla_{\bb^{t-1}}^{2} \xi_{i} 
    = \text{diag}( \hb_{i}^{(t-1)'} ) \Wb^{tT} \nabla_{\bb^{t}}^{2} \xi_{i} \Wb^{t} \text{diag}( \hb_{i}^{(t-1)'} ) 
       + \text{diag}(\hb_{i}^{(t-1)''}  \odot (\Wb^{tT} \nabla_{\bb^{t}} \xi_{i})),
  \end{equation*}
  where $t = k, \ldots, 1$.

\item $\nabla_{\Wb^{t}}^{2} \xi_{i} = ( \hb_{i}^{t-1} \otimes \hb_{i}^{(t-1)T} ) \otimes \nabla_{\bb^{t}}^{2} \xi_{i}, t = k , \ldots, 1$
  First, we have to give the order of the derivative for the Hessian matrix because the gradient of $\Wb^{t}$ is a matrix.  Thus, we define
  \[
    \nabla_{\Wb^{t}}^{2} \xi_{i} = \frac{\partial \text{Vec}( {\nabla_{\Wb^{t}} \xi_{i}} )}{\partial \text{Vec}(\Wb^{t}) }. 
  \]
  Next, we derive one block of this Hessian matrix.
  \begin{enumerate}[{Case} 1:]
  \item $t = k$.
  \begin{equation*}
  \begin{split}    
      \frac{\partial^{2} \xi_{i}}{\partial \left[ \Wb^{k} \right]_{\cdot s} \partial \left[ \Wb^{k} \right]_{\cdot r} }
    =& \frac{\partial (\left[ \nabla_{\Wb^{k}} \xi_{i} \right]_{\cdot r} ) }{\partial \left[ \Wb^{k} \right]_{\cdot s} } 
    = \frac{\partial (\left[ \nabla_{\bb^{k}} \xi_{i} \otimes \hb_{i}^{(k-1)T}  \right]_{\cdot r} ) }
            {\partial \left[ \Wb^{k} \right]_{\cdot s} } \\
    =& \frac{\partial (\left[ \hb_{i}^{(k-1)} \right]_{r} \nabla_{\bb^{k}} \xi_{i} ) }{\partial \left[ \Wb^{k} \right]_{\cdot s} } 
    = \left[ \hb_{i}^{k-1} \right]_{r} \frac{\partial \nabla_{\bb^{k}} \xi_{i} }{\partial \left[ \Wb^{k} \right]_{\cdot s} } \\
    =& \left[ \hb_{i}^{k-1} \right]_{r} 
       \left[ \frac{\partial \nabla_{\bb^{k}} \xi_{i} }{\partial \left[ \Wb^{k} \right]_{1 s} } \cdots
              \frac{\partial \nabla_{\bb^{k}} \xi_{i} }{\partial \left[ \Wb^{k} \right]_{n_{k} s} }  \right]. \\
  \end{split}
  \end{equation*}
  Now we focus on the term $\frac{\partial \nabla_{\bb^{k}} \xi_{i} }{\partial \left[ \Wb^{k} \right]_{q s} }$ with a give index $q$.
  \begin{equation*}
  \begin{split}    
       \frac{\partial \nabla_{\bb^{k}} \xi_{i} }{\partial \left[ \Wb^{k} \right]_{q s} } 
    =& \sum_{u} \frac{\partial \nabla_{\bb^{k}} \xi_{i} }{\partial \left[ \hb_{i}^{k} \right]_{u} }
                \frac{\partial \left[ \hb_{i}^{k} \right]_{u} }{\partial \left[ \Wb^{k} \right]_{q s} } 
    =  \sum_{u} \frac{\partial \nabla_{\hb^{k}} \xi_{i} }{\partial \left[ \hb_{i}^{k} \right]_{u} }
                \frac{\partial \left[ \hb_{i}^{k} \right]_{u} }{\partial \left[ \Wb^{k} \right]_{q s} } \\
    =& \sum_{u} \frac{\partial \nabla_{\hb^{k}} \xi_{i} }{\partial \left[ \hb_{i}^{k} \right]_{u} }
                \frac{\partial (\left[ \Wb^{k} \right]_{u \cdot} \hb_{i}^{k-1} + \left[ \bb^{k} \right]_{u}) }{\partial \left[ \Wb^{k} \right]_{q s} } \\
    =& \frac{\partial \nabla_{\hb^{k}} \xi_{i} }{\partial \left[ \hb_{i}^{k} \right]_{q} }
                \frac{\partial (\left[ \Wb^{k} \right]_{q \cdot} \hb_{i}^{k-1} + \left[ \bb^{k} \right]_{q}) }{\partial \left[ \Wb^{k} \right]_{q s} } 
    = \left[ \nabla_{\hb^{k}}^{2} \xi_{i} \right]_{q \cdot} \left[ \hb_{i}^{k-1} \right]_{s} \\
    =& \left[ \hb_{i}^{k-1} \right]_{s} \left[ \nabla_{\hb^{k}}^{2} \xi_{i} \right]_{\cdot q}.\\
  \end{split}
  \end{equation*}    
  We combine the aforementioned equations and we obtain
  \begin{equation*}
  \begin{split}    
      \frac{\partial^{2} \xi_{i}}{\partial \left[ \Wb^{k} \right]_{\cdot s} \partial \left[ \Wb^{k} \right]_{\cdot r} }
    =& \left[ \hb_{i}^{k-1} \right]_{r} 
       \left[ \frac{\partial \nabla_{\bb^{k}} \xi_{i} }{\partial \left[ \Wb^{k} \right]_{1 s} } \cdots
              \frac{\partial \nabla_{\bb^{k}} \xi_{i} }{\partial \left[ \Wb^{k} \right]_{n_{k} s} }  \right] \\
    =& \left[ \hb_{i}^{k-1} \right]_{r} \left[ \hb_{i}^{k-1} \right]_{s}  \nabla_{\hb^{k}}^{2} \xi_{i} 
    = \left[ \hb_{i}^{k-1} \right]_{r} \left[ \hb_{i}^{k-1} \right]_{s}  \nabla_{\bb^{k}}^{2} \xi_{i}. \\
  \end{split}
  \end{equation*}
  Therefore, we have
  \begin{equation*}
    \nabla_{\Wb^{k}}^{2} \xi_{i}
    = \frac{\partial^{2} \xi_{i}}{\partial \text{Vec}(\Wb^{k}) \partial \text{Vec}(\Wb^{k}) } 
    = (\hb_{i}^{k-1} \otimes \hb_{i}^{(k-1)T}) \otimes \nabla_{\bb^{k}}^{2} \xi_{i}. 
  \end{equation*}

  \item $t < k$. Similarly,
  \begin{equation*}
      \frac{\partial^{2} \xi_{i}}{\partial \left[ \Wb^{t} \right]_{\cdot s} \partial \left[ \Wb^{t} \right]_{\cdot r} }
    = \left[ \hb_{i}^{t-1} \right]_{r} 
      \left[ \frac{\partial \nabla_{\bb^{t}} \xi_{i} }{\partial \left[ \Wb^{t} \right]_{1 s} } \cdots
             \frac{\partial \nabla_{\bb^{t}} \xi_{i} }{\partial \left[ \Wb^{t} \right]_{n_{t} s} }  \right] .
  \end{equation*}    
  Then, we focus on the term ${\partial \nabla_{\bb^{t}} \xi_{i} }/{\partial \left[ \Wb^{t} \right]_{q s} }$ with a give index $q$.
  \begin{equation*}
  \begin{split}    
      \frac{\partial \nabla_{\bb^{t}} \xi_{i} }{\partial \left[ \Wb^{t} \right]_{q s} } 
    =& \sum_{u} \frac{\partial \nabla_{\bb^{t}} \xi_{i} }{\partial \left[ \hb_{i}^{t} \right]_{u} }
                \frac{\partial \left[ \hb_{i}^{t} \right]_{u} }{\partial \left[ \Wb^{t} \right]_{q s} } 
    = \sum_{u} \frac{\partial \nabla_{\bb^{t}} \xi_{i} }{\partial \left[ \hb_{i}^{t} \right]_{u} }
                \frac{\partial \sigma(\left[ \Wb^{t} \right]_{u \cdot} \hb_{i}^{t-1} + \left[ \bb^{t} \right]_{u}) }
                     {\partial \left[ \Wb^{t} \right]_{q s} } \\
    =& \frac{\partial \nabla_{\bb^{t}} \xi_{i} }{\partial \left[ \hb_{i}^{t} \right]_{q} }
       \frac{\partial \sigma(\left[ \Wb^{t} \right]_{q \cdot} \hb_{i}^{t-1} + \left[ \bb^{t} \right]_{q}) }
            {\partial \left[ \Wb^{t} \right]_{q s} } \\
    =& \frac{\partial \nabla_{\bb^{t}} \xi_{i} }{\partial \left[ \hb_{i}^{t} \right]_{q} }
       \sigma'(\left[ \Wb^{t} \right]_{q \cdot} \hb_{i}^{t-1} + \left[ \bb^{t} \right]_{q}) \left[ \hb_{i}^{t-i} \right]_{s}  
    =  \left[ \hb_{i}^{t-i} \right]_{s} \frac{\partial \nabla_{\bb^{t}} \xi_{i} }{\partial \left[ \hb_{i}^{t} \right]_{q} }
       \frac{\partial \left[ \hb_{i}^{t} \right]_{q} }{\partial \left[ \bb^{t} \right]_{q} } \\
    =& \left[ \hb_{i}^{t-i} \right]_{s} 
       \sum_{u} \frac{\partial \nabla_{\bb^{t}} \xi_{i} }{\partial \left[ \hb_{i}^{t} \right]_{u} }
                \frac{\partial \left[ \hb_{i}^{t} \right]_{u} }{\partial \left[ \bb^{t} \right]_{q} } 
    = \left[ \hb_{i}^{t-i} \right]_{s} \frac{\partial \nabla_{\bb^{t}} \xi_{i} }{\partial \left[ \bb^{t} \right]_{q} } 
    = \left[ \hb_{i}^{t-i} \right]_{s} \left[ \nabla_{\bb^{t}}^{2} \xi_{i} \right]_{q \cdot}  \\
    =& \left[ \hb_{i}^{t-i} \right]_{s} \left[ \nabla_{\bb^{t}}^{2} \xi_{i} \right]_{\cdot q}.  \\
  \end{split}
  \end{equation*}    
  We combine the aforementioned equations and obtain
  \begin{equation*}
  \begin{split}    
       \frac{\partial^{2} \xi_{i}}{\partial \left[ \Wb^{t} \right]_{\cdot s} \partial \left[ \Wb^{t} \right]_{\cdot r} }
    =& \left[ \hb_{i}^{t-1} \right]_{r} 
       \left[ \frac{\partial \nabla_{\bb^{t}} \xi_{i} }{\partial \left[ \Wb^{t} \right]_{1 s} } \cdots
              \frac{\partial \nabla_{\bb^{t}} \xi_{i} }{\partial \left[ \Wb^{t} \right]_{n_{t} s} }  \right] \\
    =& \left[ \hb_{i}^{t-1} \right]_{r} \left[ \hb_{i}^{t-1} \right]_{s}  \nabla_{\bb^{t}}^{2} \xi_{i}. \\
  \end{split}
  \end{equation*}
  Therefore, we have
  \begin{equation*}
  \begin{split}
    \nabla_{\Wb^{t}}^{2} \xi_{i}
    = \frac{\partial^{2} \xi_{i}}{\partial \text{Vec}(\Wb^{t}) \partial \text{Vec}(\Wb^{t}) } 
    = (\hb_{i}^{t-1} \otimes \hb_{i}^{(t-1)T}) \otimes \nabla_{\bb^{t}}^{2} \xi_{i}. 
  \end{split}
  \end{equation*}

\end{enumerate}

\end{enumerate}

\clearpage
\section{Supplementary Materials D}
The difference between original Hessian and expectation approximate Hessian in the $(t-1)^{\text{th}}$ layer is shown as 
\begin{align}
   & \Expectation_{i}[\text{diag}(\hb_{i}^{(t-1)'}) \Wb^{tT} \nabla_{\bb^{t}}^{2} \xi_{i} \Wb^{t} \text{diag}(\hb_{i}^{(t-1)'}) 
     + \text{diag}(\hb_{i}^{(t-1)''}  \odot (\Wb^{tT} \nabla_{\bb^{t}} \xi_{i}))] \nonumber \\
   & - \Expectation_{i}[\text{diag}(\hb_{i}^{(t-1)'}) \Wb^{tT} \Expectation_{i}[ \nabla_{\bb^{t}}^{2} \xi_{i}] \Wb^{t} \text{diag}(\hb_{i}^{(t-1)'}) 
     + \text{diag}(\hb_{i}^{(t-1)''}  \odot (\Wb^{tT} \nabla_{\bb^{t}} \xi_{i}))] \nonumber \\
  =& \Expectation_{i}[ (\Wb^{tT} \nabla_{\bb^{t}}^{2} \xi_{i} \Wb^{t}) \odot (\hb_{i}^{(t-1)'} \otimes \hb_{i}^{(t-1)'T} )] - (\Wb^{tT} \Expectation_{i}[ \nabla_{\bb^{t}}^{2} \xi_{i}] \Wb^{t}) \odot \text{EhhT}^{(t-1)'} \nonumber \\
  =& \text{Ele-Cov}( \Wb^{tT} \nabla_{\bb^{t}}^{2} \xi_{i} \Wb^{t}, \hb_{i}^{(t-1)'} \otimes \hb_{i}^{(t-1)'T} ). \label{eq:Hessian_Diff}
\end{align}
To find the bound of Eq.~(\ref{eq:Hessian_Diff}), we utilize the inequality of covariance
\[
  \text{Cov}(X, Y)^{2} \leq \text{Var}(X) \text{Var}(Y)
\]
which is implied by the Cauchy-Schwarz inequality. Thus, given indices $\mu$ and $\nu$, we have
\[
       \text{Cov}( [\Wb^{tT} \nabla_{\bb^{t}}^{2} \xi_{i} \Wb^{t}]_{\mu \nu}, [\hb_{i}^{(t-1)'} \otimes \hb_{i}^{(t-1)'T}]_{\mu \nu} )^{2} 
  \leq \text{Var}([\Wb^{tT} \nabla_{\bb^{t}}^{2} \xi_{i} \Wb^{t}]_{\mu \nu}) \text{Var}([\hb_{i}^{(t-1)'} \otimes \hb_{i}^{(t-1)'T}]_{\mu \nu}).
\]
Therefore, 
\begin{align}
      & \left\| \text{Ele-Cov}( \Wb^{tT} \nabla_{\bb^{t}}^{2} \xi_{i} \Wb^{t}, \hb_{i}^{(t-1)'} \otimes \hb_{i}^{(t-1)'T} ) \right\|_{F}^{2} \nonumber \\
     =& \sum_{\mu, \nu} \text{Cov}( [\Wb^{tT} \nabla_{\bb^{t}}^{2} \xi_{i} \Wb^{t}]_{\mu \nu}, 
                                    [\hb_{i}^{(t-1)'} \otimes \hb_{i}^{(t-1)'T}]_{\mu \nu} )^{2} \nonumber \\
  \leq& \sum_{\mu, \nu} \text{Var}([\Wb^{tT} \nabla_{\bb^{t}}^{2} \xi_{i} \Wb^{t}]_{\mu \nu}) 
                        \text{Var}([\hb_{i}^{(t-1)'} \otimes \hb_{i}^{(t-1)'T}]_{\mu \nu}). \label{eq:Hessian_Error_Bound}
\end{align}
Obviously, $\text{Var}([\hb_{i}^{(t-1)'} \otimes \hb_{i}^{(t-1)'T}]_{\mu \nu})$ can be controlled easier than $\text{Var}([\Wb^{tT} \nabla_{\bb^{t}}^{2} \xi_{i} \Wb^{t}]_{\mu \nu})$. At the beginning, we suppose the activation function $\sigma$ is $L$-Lipschitz continuous, i.e.,
\[
  |\sigma(x) - \sigma(y)| \leq L |x-y|,
\]
which implies 
\[
  |\sigma'(x)| \leq L.
\]
Thus, 
\begin{align}
         \text{Var}([\hb_{i}^{(t-1)'} \otimes \hb_{i}^{(t-1)'T}]_{\mu \nu}) 
     = & \Expectation_{i}[ [\hb_{i}^{(t-1)'} \otimes \hb_{i}^{(t-1)'T}]_{\mu \nu}^{2} ] 
         - \Expectation_{i}[ [\hb_{i}^{(t-1)'} \otimes \hb_{i}^{(t-1)'T}]_{\mu \nu} ]^{2} \nonumber \\
     = & \Expectation_{i}[ [\hb_{i}^{(t-1)'}]_{\mu}^{2} [\hb_{i}^{(t-1)'}]_{\nu}^{2} ] 
         - \Expectation_{i}[ [\hb_{i}^{(t-1)'}]_{\mu} [\hb_{i}^{(t-1)'}]_{\nu} ]^{2}  \nonumber \\
  \leq & \Expectation_{i}[ [\hb_{i}^{(t-1)'}]_{\mu}^{2} [\hb_{i}^{(t-1)'}]_{\nu}^{2} ]  \nonumber \\
  \leq & \Expectation_{i}[ L^{2} L^{2} ]  = L^{4}. \label{eq:Activation_Variance_Bound} 
\end{align}
At the final, we combine Eq.~(\ref{eq:Hessian_Error_Bound}) and Eq.~(\ref{eq:Activation_Variance_Bound}) to
\begin{equation*}
\label{eq:Hessian_Error_Bound_with_L_Lipschitz}
  \left\| \text{Ele-Cov}( \Wb^{tT} \nabla_{\bb^{t}}^{2} \xi_{i} \Wb^{t}, \hb_{i}^{(t-1)'} \otimes \hb_{i}^{(t-1)'T} ) \right\|_{F}^{2} 
  \leq  L^{4} \sum_{\mu, \nu} \text{Var}([\Wb^{tT} \nabla_{\bb^{t}}^{2} \xi_{i} \Wb^{t}]_{\mu \nu}),
\end{equation*}
and the proof is done.

We give an example here, the sigmoid function is 0.25-Lipschitz. That is, $L_{\text{sigmoid}} = 0.25$. Hence, when we utilized the sigmoid as our activation function in FCNNs, we have the bound 
\[
  0.00390625 \cdot \sum_{\mu, \nu} \text{Var}([\Wb^{tT} \nabla_{\bb^{t}}^{2} \xi_{i} \Wb^{t}]_{\mu \nu})
\]
for the errors from the backpropagation of the second derivatives in the $(t-1)^\text{th}$ layer.

\end{document}